\documentclass{article}

\usepackage[final]{neurips_data_2023}

\usepackage[utf8]{inputenc} 
\usepackage[T1]{fontenc}    
\usepackage{hyperref}       
\usepackage{url}            
\usepackage{booktabs}       
\usepackage{amsfonts}       
\usepackage{nicefrac}       
\usepackage{microtype}      
\usepackage{xcolor}         
\usepackage{multirow}       
\usepackage{graphicx}
\usepackage{enumitem}

\newcommand{\eg}{\textit{e}.\textit{g}.}
\newcommand{\ie}{\textit{i}.\textit{e}.}
\newcommand{\etal}{\textit{et}.\textit{al}.}

\title{When Super-Resolution Meets Camouflaged Object Detection: A Comparison Study}

\author{%
Juan Wen$^{1,2}$,
 Shupeng Cheng$^{3}$,
 Peng Xu$^{3}$,
Bowen Zhou$^{3}$, 
{\bf Radu Timofte$^{2,4}$}\, \\
{\bf Weiyan Hou$^{2}$\thanks{Corresponding authors.}  , }   
{   \bf Luc Van Gool$^{2}$}\\
  $~^{1}$Zhengzhou University\\
  $~^{2}$Computer Vision Lab,
 ETH Zurich\\
  $~^{3}$Tsinghua University\\
  $~^{4}$University of Wurzburg  \\
  \texttt{\{juawen,timofter,vangool\}@vision.ee.ethz.ch },
  \texttt{houwy@zzu.edu.cn} \\
  \texttt{\{peng\_xu,zhoubowen\}@tsinghua.edu.cn },
  \texttt{chengsp20@mails.tsinghua.edu.cn}
}

\begin{document}

\maketitle

\begin{abstract}
Super Resolution (SR) and Camouflaged Object Detection (COD) are two hot topics in computer vision with various joint applications.
For instance, low-resolution surveillance images can be successively processed by super-resolution techniques and camouflaged object detection.
However, in previous work, these two areas are always studied in isolation. In this paper, we, for the first time, conduct an integrated comparative evaluation for both. 
Specifically, we benchmark different super-resolution methods on commonly used COD datasets, and meanwhile, we evaluate the robustness of different COD models by using COD data processed by SR methods.
Our goal is to bridge these two domains, discover novel experimental phenomena, summarize new experimental regularities, and provide insights and inspiration for future research work in both communities. 

\end{abstract}

\section{Introduction}   

Super-resolution(SR) is a technique employed to reconstruct visually pleasing, high-quality images from blurry, noisy, or low-resolution inputs \cite{zhang2021plug}. Camouflaged Object Detection (COD) encompasses the challenging task of accurately detecting and localizing camouflaged objects within images captured in diverse natural and artificial environments. As two prominent research directions, SR and COD share numerous joint application scenarios, including critical areas such as security surveillance and reconnaissance, remote sensing image analysis, as well as advanced medical image processing analysis and autonomous cars.\\ \\
However, in previous work, those two areas were always studied in isolation. In this paper, we present an integrated comparative evaluation of both, which, to the best of our knowledge, is the first of its kind. Specifically, on the one hand, we evaluate the performance of five super-resolution methods on four commonly used datasets in the camouflaged object detection domain. On the other hand, we evaluate the performance of nine COD models by feeding them with COD datasets processed by five different SR methods, which allows us to assess both the robustness of the COD models and view the performance of the SR methods from a novel perspective.\\  \\
In our experiments, we observed variations in the performance of the five SR methods on the COD datasets compared to their performance on traditional SR datasets. Besides, we found that when COD datasets undergo downsampling and super-resolution processing, DGNet \cite{ji2023deep}exhibits the best overall performance among the nine models. Further experimental details and observations  will be discussed in Section 3 and 4.\\In summary, our contributions are as follows:
\begin{itemize}[label=$\bullet$, align=left, leftmargin=*]
\item We innovatively integrate the fields of super-resolution (SR) and camouflage object detection (COD). 
\item We conduct comprehensive experiments using five different super-resolution methods and nine COD models, enabling a quantitative evaluation of the robustness of various COD models and a novel perspective on evaluating different SR methods.
\item We analyze the experimental data and summarize some experimental regularities, providing insights and directions for future research in both the fields of super-resolution and camouflage object detection.
\end{itemize}

\section{Related work}
\subsection{Super-Resolution}   
Super Resolution (SR) is an image processing technique aimed at enhancing low-resolution images to high-resolution levels. One commonly used method to achieve this goal is bicubic SR, which utilizes bicubic \cite{keys1981cubic} interpolation to increase image details and clarity. Bicubic \cite{keys1981cubic}SR has been widely applied in traditional image processing.
With the advancement of modern deep learning techniques, advanced super-resolution technologies have emerged. Enhanced Deep Super-Resolution EDSR\cite{lim2017enhanced}, is one such super-resolution method that leverages deep learning and is based on the concept of residual networks. Compared to traditional interpolation techniques, EDSR \cite{lim2017enhanced}  exhibits outstanding performance in enhancing image details and textures.
ESRGAN \cite{wang2018esrgan}is a deep supervised learning method for super-resolution based on generator and discriminator networks. Adversarial training between the generator and discriminator networks encourages the generator network to acquire better super-resolution reconstruction capabilities, resulting in results that are closer to real high-resolution images. It outperforms traditional methods and other deep learning-based super-resolution methods.
SwinIR \cite{liang2021swinir}, a novel image restoration technique, is based on the Swin Transformer model developed by Liu \cite{liu2021swin}. SwinIR utilizes attention mechanisms for various tasks, including image and language processing. By leveraging Swin Transformer's ability to model remote contextual information and global correlations, SwinIR improves the quality of image restoration.
More recently, Cross Aggregation Transformer (CAT) \cite{chen2022cross}and Attention Shrinkage-based Precise Image Restoration Technique (ART) \cite{zhang2022accurate}have demonstrated state-of-the-art performance in various image restoration tasks, such as image super-resolution, denoising, and removal of JPEG compression artifacts. These techniques are based on transformer-based image restoration technology, which offers superior performance compared to existing convolutional neural network-based methods due to its parameter-independent global interactions.

\subsection{Camouflaged Object Detection}   
In recent years, camouflage object detection has garnered increasing attention in the field of computer vision. Early COD methods relied on hand-crafted features, including color, texture, motion and gradient \cite{mondal2020camouflaged}. However, due to the limited expressive power of hand-crafted features, these methods often fail to achieve satisfactory results in complex scenes. With the emergence of deep learning, many CNN-based models have shown promising performance in camouflaged object detection (COD). Fan \etal~ \cite{fan2020camouflaged} utilized a multi-stage strategy, proposing first to locate and then detect camouflage objects for better performance. Yan \etal~ \cite{yan2021mirrornet}introduced MirrorNet that utilizes both instance segmentation and adversarial attack techniques for effective detection. Sun \etal~ \cite{sun2021context}presented C2FNet, incorporating an attention-induced cross-level fusion module and a dual-branch global context module to aggregate features from multiple levels and enhance feature representation. Inspired by the detection process of predators, Mei \etal ~\cite{mei2021camouflaged}proposed PFNet, which includes a positioning module and a focus module to effectively detect camouflaged objects. Zhuge \etal ~\cite{zhuge2022cubenet}designed CubeNet, which combinines attention fusion and X-shaped connections to aggregate multi-scale features. Zhai \etal ~\cite{zhai2021mutual}developed a mutual graph learning model that separates one input into different features to roughly locate the object and accurately identifying its boundary. Furthermore, Ji \etal ~\cite{ji2023deep}introduced DGNet, which excavates texture information by learning object-level gradient instead of relying on boundary-aware or uncertainty-aware modeling.


\section{Experiment}   
\subsection{Hardware Information}  
We conducted all evaluation tasks on a NVIDIA GeForce RTX 3090 GPU with 24GB of memory.

\subsection{Dataset}   
In this experiment, we selected four widely used datasets in the field of camouflage object detection (COD): CHAMELEON \cite{fan2021concealed}, CAMO \cite{le2019anabranch}, COD10K \cite{fan2021concealed}, and NCK4 \cite{lv2021simultaneously}. CHAMELEON consists of 76 images obtained through a Google search for "camouflaged animals" and is solely used for testing purposes. CAMO comprises 1250 images of camouflaged objects, with 1000 images for training and 250 images for testing. COD10K includes a collection of 10,000 images from multiple websites, covering 10 super-classes and 78 sub-classes. Among the 10,000 images, 5066 are camouflaged, with 3044 images dedicated to training and 2026 images for testing. NC4K is currently the largest dataset used to test COD models and consists of 4121 camouflaged images in natural and artificial scenes.

\subsection{Metrics}   
The evaluation of the SR models adopts two widely used evaluation indicators: SSIM (Structural SIMilarity) and PSNR (Peak Signal-to-Noise Ratio) \cite{hore2010image}.\\ \\
The evaluation of the COD models adopts four commonly used metrics, \ie, mean absolute error \textit{M}, \cite{perazzi2012saliency}, weighted F-measure $\textit{F}^{w}_{\beta}$, \cite{margolin2014evaluate}, structure-measure $\textit{S}_{\alpha}$, \cite{fan2017structure} and mean E-measure $\textit{E}_{\phi}$, \cite{fan2021cognitive}.\\ \\
Additionally, to quantitatively compare the extent of performance degradation among different COD models, we propose two new metrics: Avg\_D and DM. Specifically, when the performance of the COD model decreases, the metrics $\textit{S}_{\alpha}$, $\textit{E}_{\phi}$, and $\textit{F}^{w}_{\beta}$ decrease, while the metric \textit{M} increases. Therefore, we use the average percentage decrease of $\textit{S}_{\alpha}$, $\textit{E}_{\phi}$, and $\textit{F}^{w}_{\beta}$ as the metric Avg\_D, and the percentage increase of \textit{M} as the metric DM to serve as quantitative measures for evaluating the performance degradation of COD models.

\subsection{Evaluation of Camouflaged Object Detection Models}  
We selected nine COD models, namely SINet \cite{fan2020camouflaged}, SINet-v2 (\cite{fan2021concealed}), DGNet \cite{ji2023deep}, FAPNet [9] \cite{zhou2022feature}, C2FNet \cite{sun2021context}, C2FNet-v2 \cite{chen2022camouflaged}, BGNet \cite{sun2022boundary}, Camoformer \cite{yin2022camoformer}, and UR-COD \cite{kajiura2021improving} for evaluation. During the experiment, we fed the images processed by different super-resolution methods into these nine COD models individually and compared their output results with the ground truth (GT). The images that underwent downsampling and subsequent super-resolution modeling would lose some information compared to the original images. Consequently, inputting these images into COD models might lead to a decline in the performance of camouflage object detection. One of the aims of our experiment is to compare the extent of performance degradation among different COD models under this circumstance. This serves as a novel perspective to evaluate the robustness of COD models.

\subsection{Evaluation of Super-Resolution Techniques}  
Initially, we evaluated different SR methods on the four COD datasets mentioned above. Specifically, we utilized the BICUBIC \cite{keys1981cubic} 4x down-sampling method in Matlab to obtain low-quality images. Then, we applied five different 4x SR methods, namely EDSR \cite{lim2017enhanced} ESRGAN \cite{wang2018esrgan}, SwinIR \cite{liang2021swinir}, ART \cite{zhang2022accurate}, and CAT \cite{chen2022cross} to the low-quality images and compared the output images with the ground truth (GT) using the SSIM (Structural Similarity) and PSNR (Peak Signal-to-Noise Ratio) \cite{hore2010image} metrics. \\
Furthermore, the aforementioned experiments of subjecting the datasets to downsampling, SR methods, and COD models in sequence provide a novel perspective for evaluating the performance of SR methods.




\section{Results and analysis}










\subsection{Results and Analysis on the Evaluation of Different Super-Resolution Methods}
After applying five SR methods to four common COD datasets:
\begin{itemize}[label=$\bullet$, align=left, leftmargin=*]
\item We found that the CAT \cite{chen2022cross} (Cross Aggregation Transformer) model achieved the best performance among the five SR methods, as it is currently the state-of-the-art method. This indicates that the CAT \cite{chen2022cross}model utilizes Rectangle-Window Self-Attention (Rwin-SA) to expand the attention area and aggregate features across different windows using horizontal and vertical rectangular window attention with different heads. It incorporates Axial-Shift operation for interactions between different windows. The Locality Complementary Module is used to complement the self-attention mechanism, combining the inductive biases of CNN(such as translation invariance and locality) into the Transformer, thus achieving global-local coupling in this image restoration model, which exhibits good generalization performance in real-world scenarios. However, in our experiments, we found that the CAT \cite{chen2022cross} model has a deep, large, and complex network structure, requiring more resources, time, and expertise for training and testing.

\item The ART \cite{zhang2022accurate} model is the second most advanced method among the five SR methods, and it also performs well, achieving a sub-optimal performance on the four common COD datasets.

\item EDSR \cite{lim2017enhanced}is the champion of the NTIRE 2017 SR \cite{agustsson2017ntire} challenge. It is based on SRResNet \cite{ledig2017photo} but enhances the network by removing normalization layers and using a deeper and wider network structure. The EDSR \cite{lim2017enhanced} model demonstrates excellent performance in terms of Peak Signal-to-Noise Ratio (PSNR) and Structural Similarity Index (SSIM ) on the four common COD datasets. It even surpasses SwinIR \cite{liang2021swinir} and ESRGAN \cite{wang2018esrgan}, indicating good real-world generalization performance of the EDSR \cite{lim2017enhanced} method.

\item In the four COD datasets, when testing with models trained using SwinIR \cite{liang2021swinir} and EDSR \cite{lim2017enhanced}, we observed that SwinIR achieved lower SSIM (Structural SIMilarity) and PSNR (Peak Signal-to-Noise Ratio) \cite{hore2010image} results compared to EDSR \cite{lim2017enhanced}. The potential reasons for this are as follows:

Differences in training datasets: SwinIR and EDSR \cite{lim2017enhanced} use different training datasets. The diversity and representativeness of the training dataset are crucial for obtaining good results on real-world test images. If SwinIR's \cite{liang2021swinir} training dataset does not match the features and content of the test images, it may lead to degraded performance.

Selection of model architecture and parameters: SwinIR \cite{liang2021swinir} and EDSR \cite{lim2017enhanced} have different network architectures and parameter settings. On the same real-world image dataset, SwinIR's \cite{liang2021swinir} model architecture and parameter settings may not be as suitable, resulting in lower performance compared to EDSR \cite{lim2017enhanced}. The choice of model architecture and parameters can be critical for achieving better results on specific datasets.

Training process and strategies: SwinIR \cite{liang2021swinir} and EDSR \cite{lim2017enhanced} also differ in their training process and strategies. Different training strategies can lead to different model performances on real-world images. Factors such as data augmentation, choice of loss functions, and hyperparameters during training can all influence the model's performance.

Image content and characteristics: The content and characteristics of real-world image datasets can also impact the results of super-resolution methods. Some images may be better suited for EDSR's \cite{lim2017enhanced} approach in capturing details and textures, while others may be more suitable for SwinIR's \cite{liang2021swinir} approach. Therefore, different characteristics of different images may lead to lower performance of SwinIR \cite{liang2021swinir} on certain images. Overall, this suggests that SwinIR has a slightly weaker generalization performance than EDSR \cite{lim2017enhanced} in real-world scenarios.

\item ESRGAN performed the worst in terms of  SSIM (Structural SIMilarity) and PSNR (Peak Signal-to-Noise Ratio) \cite{hore2010image} among the five SR methods tested on the four common COD datasets. However, it should be noted that ESRGAN is a GAN-based method introduced in ECCV 2018 \cite{wang2018esrgan}, and its design primarily focuses on generating high-quality images rather than optimizing for PSNR.
\end{itemize}

    
\begin{table}
    \caption{\textbf{Results of PSNR and SSIM test metrics for five SR methods on four COD test datasets}}
    \centering
    \resizebox{\textwidth}{!}{
    \begin{tabular}{*{10}{c}}
        \toprule
        \multirow{2}{*}{Method} & \multirow{2}{*}{Scale} & \multicolumn{2}{c}{CAMO} & \multicolumn{2}{c}{COD10K} & \multicolumn{2}{c}{NC4K} & \multicolumn{2}{c}{CHAMELEON} \\
        \cmidrule(lr){3-4} \cmidrule(lr){5-6} \cmidrule(lr){7-8} \cmidrule(lr){9-10}
         &  & $\textit{PSNR}_\uparrow$ & $\textit{SSIM}_\uparrow$ & $\textit{PSNR}_\uparrow$ & $\textit{SSIM}_\uparrow$ & $\textit{PSNR}_\uparrow$ & $\textit{SSIM}_\uparrow$ & $\textit{PSNR}_\uparrow$ & $\textit{SSIM}_\uparrow$ \\
        \midrule
        EDSR & X4 & 22.54 & 0.5738 & 24.93 & 0.6670 & 24.13 & 0.6354 & 22.02 & 0.5416 \\
        ESRGAN & X4 & 21.12 & 0.5140 & 23.48 & 0.6147 & 22.76 & 0.5843 & 20.72 & 0.4856 \\
        SwinIR (classical-SR) & X4 & 22.46 & 0.5700 & 24.85 & 0.6637 & 24.05 & 0.6321 & 21.95 & 0.5393 \\
        ART & X4 & 24.40 & 0.6161 & 26.96 & 0.7110 & 26.15 & 0.6808 & 23.80 & 0.5783 \\
        CAT & X4 & 24.45 & 0.6180 & 27.00 & 0.7126 & 26.18 & 0.6823 & 23.83 & 0.5796 \\
        \bottomrule
    \end{tabular}
    }
\end{table}

\subsection{Results and Analysis on the Evaluation of Different Camouflaged Object Detection Models}   
We quantitatively evaluated the performance degradation of nine COD models when presented with datasets processed by five different SR methods using the two new metrics mentioned earlier. The results are depicted in Figures 1 to 8. \\
\begin{figure}[htbp]
  \centering
  \includegraphics[width=\textwidth]{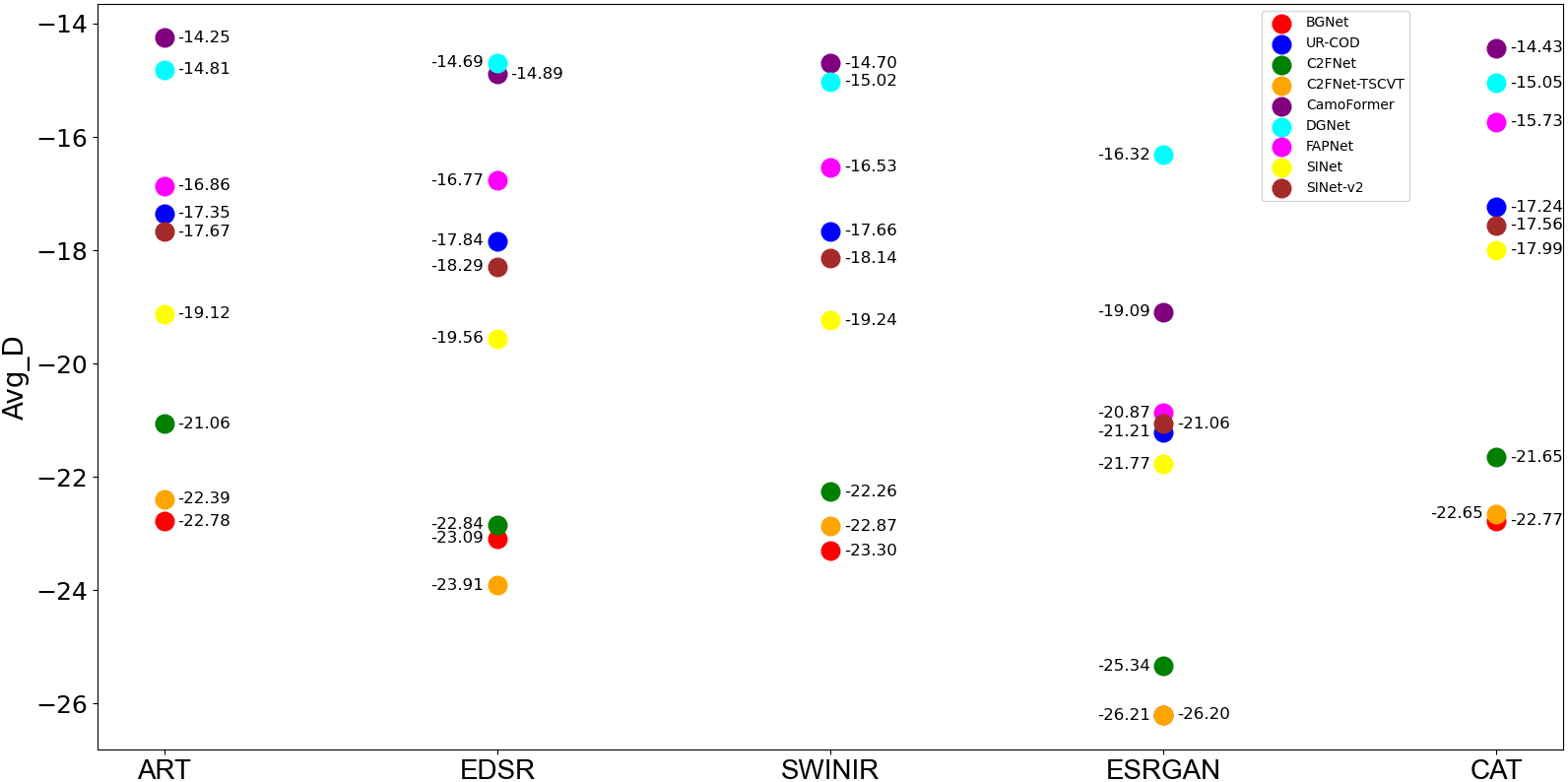} 
  \caption{scatter plot of Avg\_D on CAMO}
\end{figure}

\begin{figure}[htbp]
  \centering
  \includegraphics[width=\textwidth]{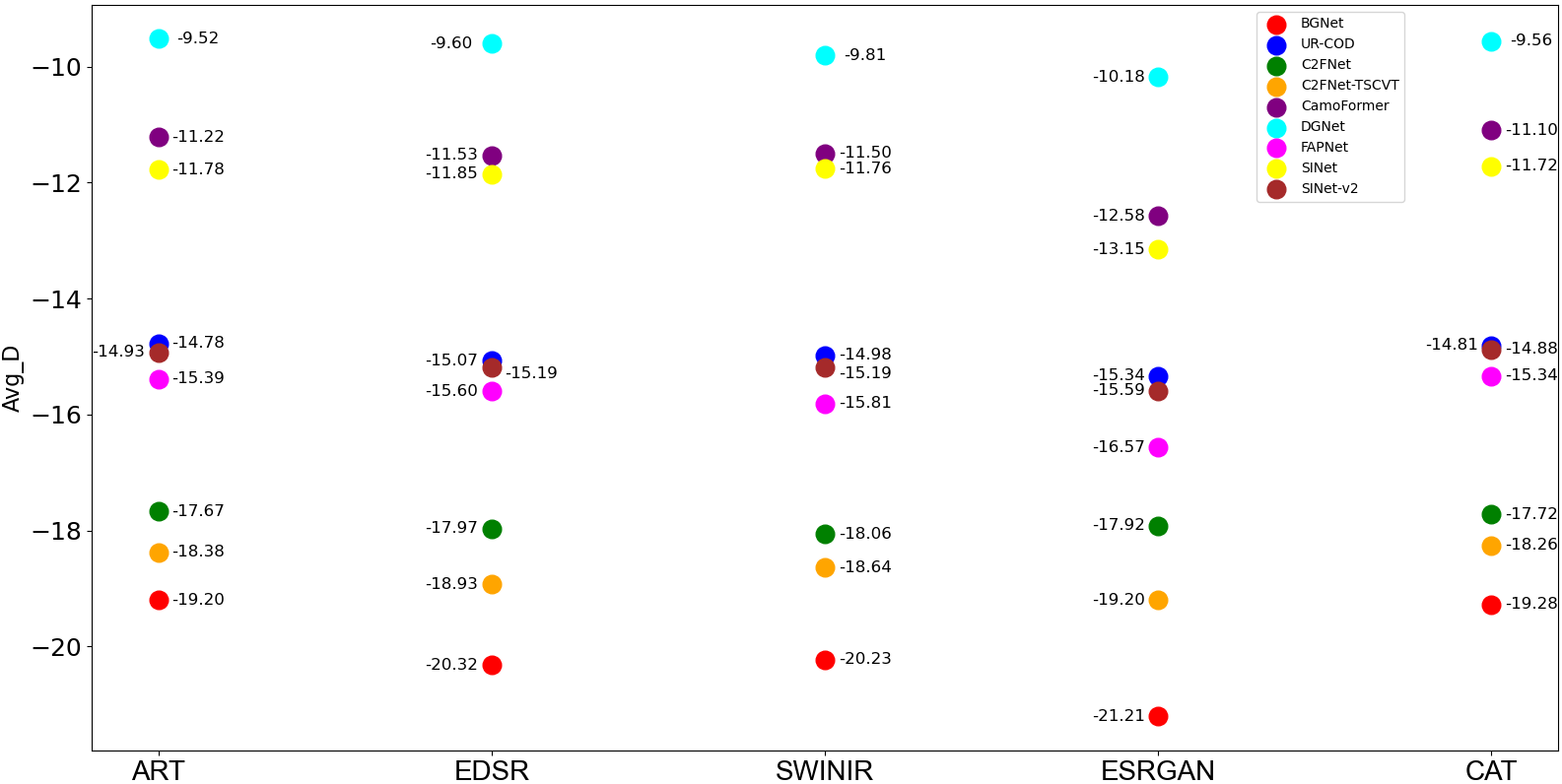} 
  \caption{scatter plot of Avg\_D on COD10K}
\end{figure}

\begin{figure}[htbp]
  \centering
  \includegraphics[width=\textwidth]{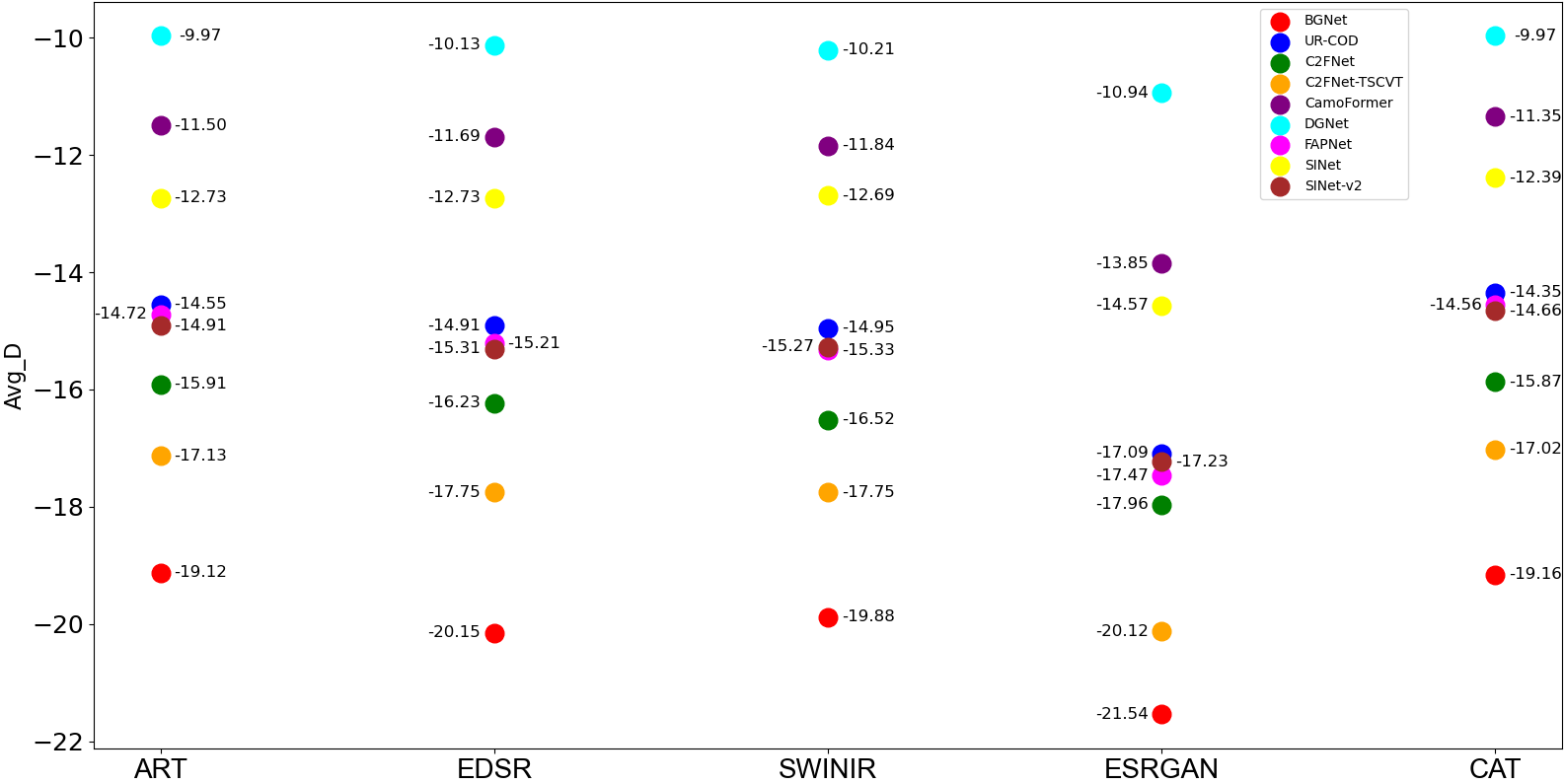} 
  \caption{scatter plot of Avg\_D on NC4K}
\end{figure}

\begin{figure}[htbp]
  \centering
  \includegraphics[width=\textwidth]{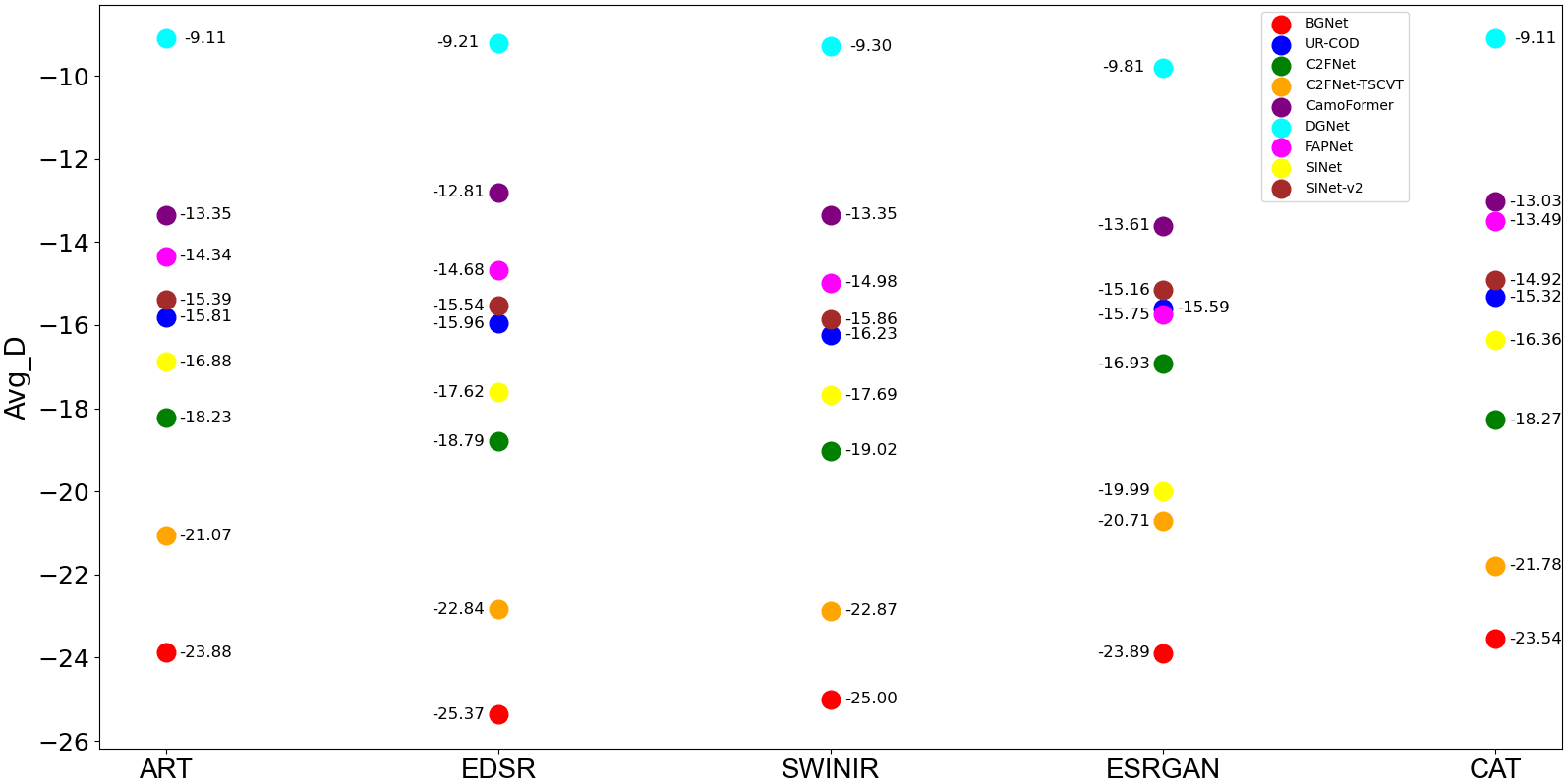} 
  \caption{scatter plot of Avg\_D on CHAMELEON}
\end{figure}

\begin{figure}[htbp]
  \centering
  \includegraphics[width=\textwidth]{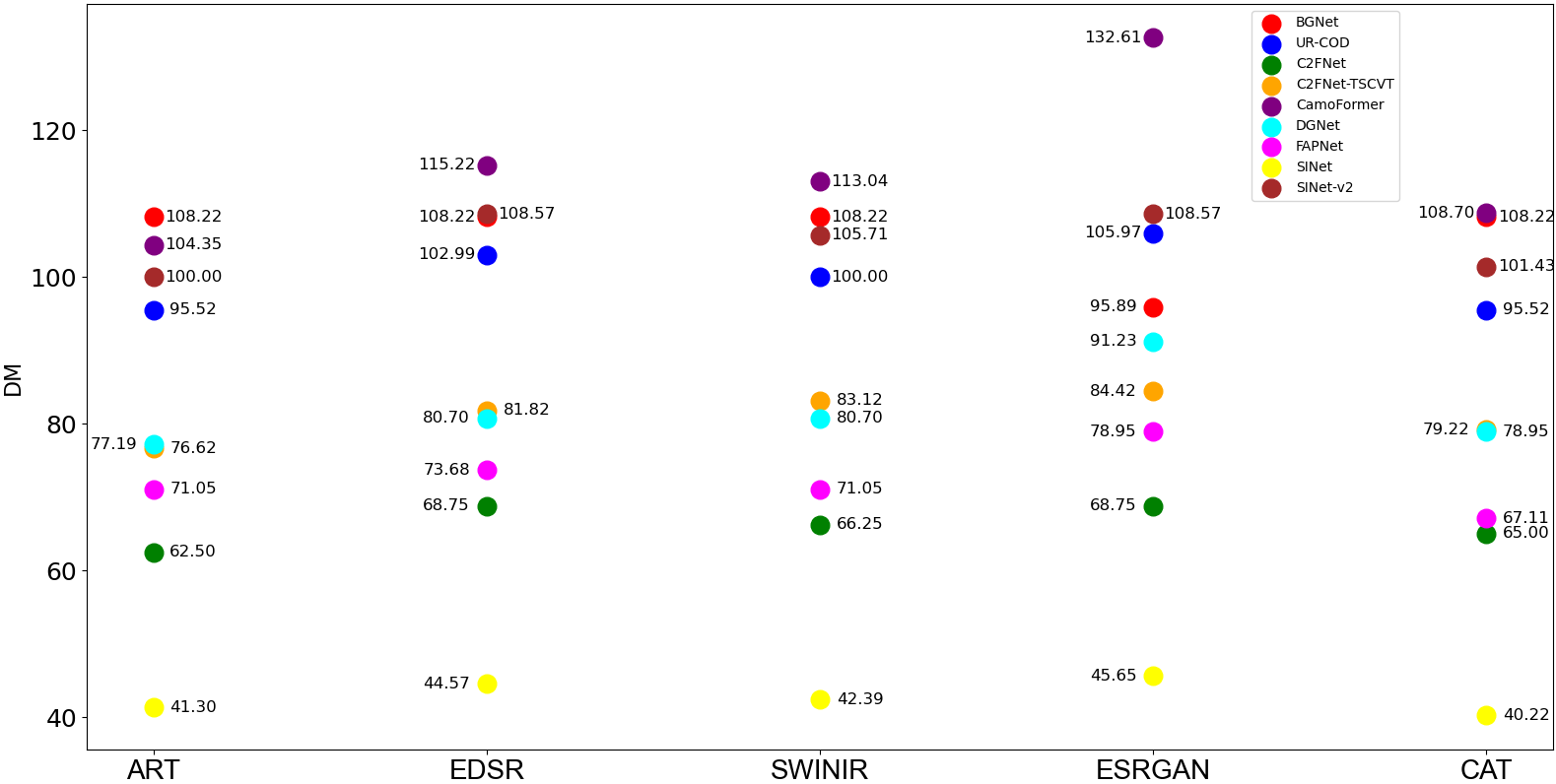} 
  \caption{scatter plot of DM on CAMO}
\end{figure}

\begin{figure}[htbp]
  \centering
  \includegraphics[width=\textwidth]{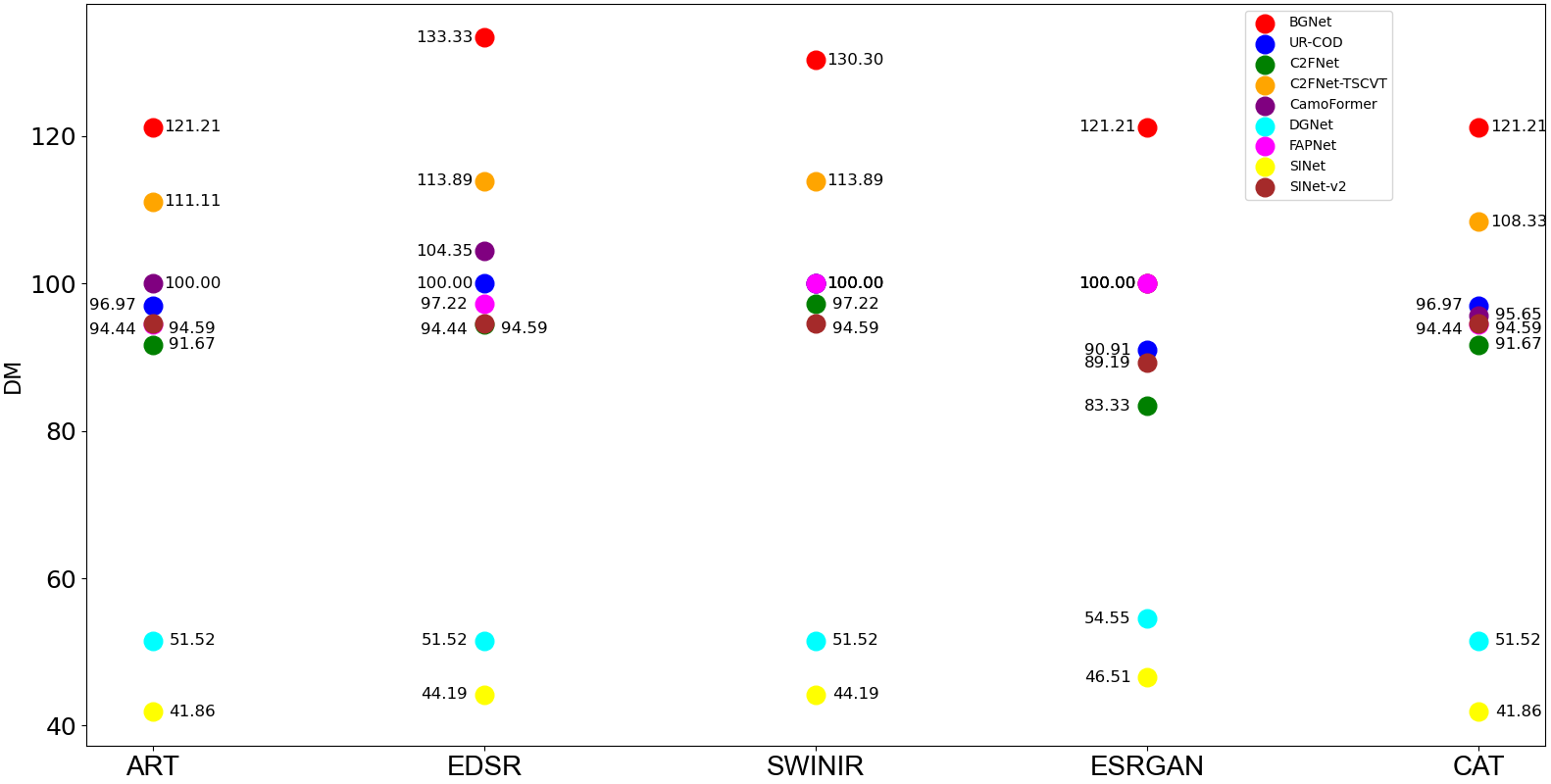} 
  \caption{scatter plot of DM on COD10K}
\end{figure}

\begin{figure}[htbp]
  \centering
  \includegraphics[width=\textwidth]{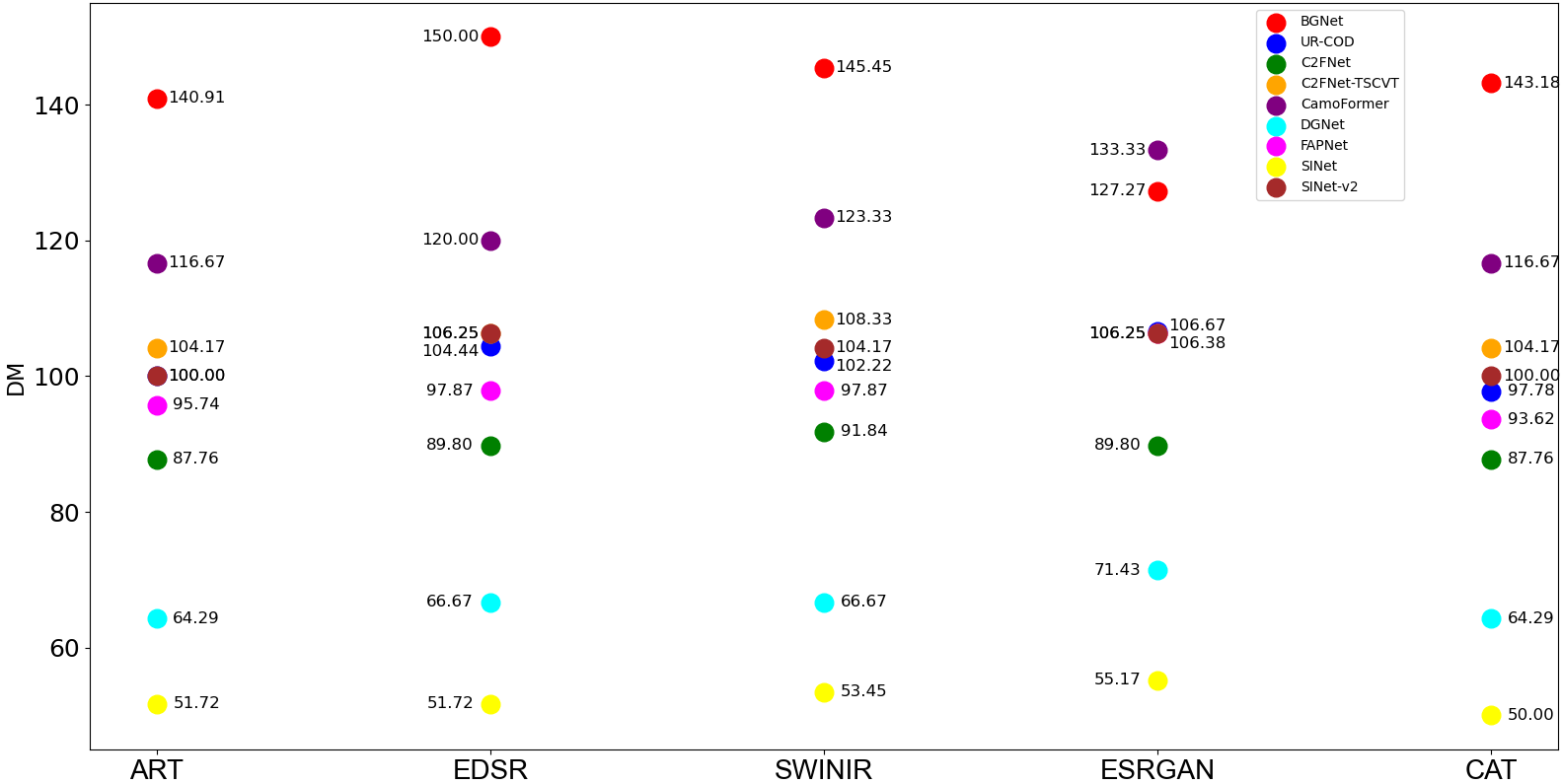} 
  \caption{scatter plot of DM on NC4K}
\end{figure}

\begin{figure}[htbp]
  \centering
  \includegraphics[width=\textwidth]{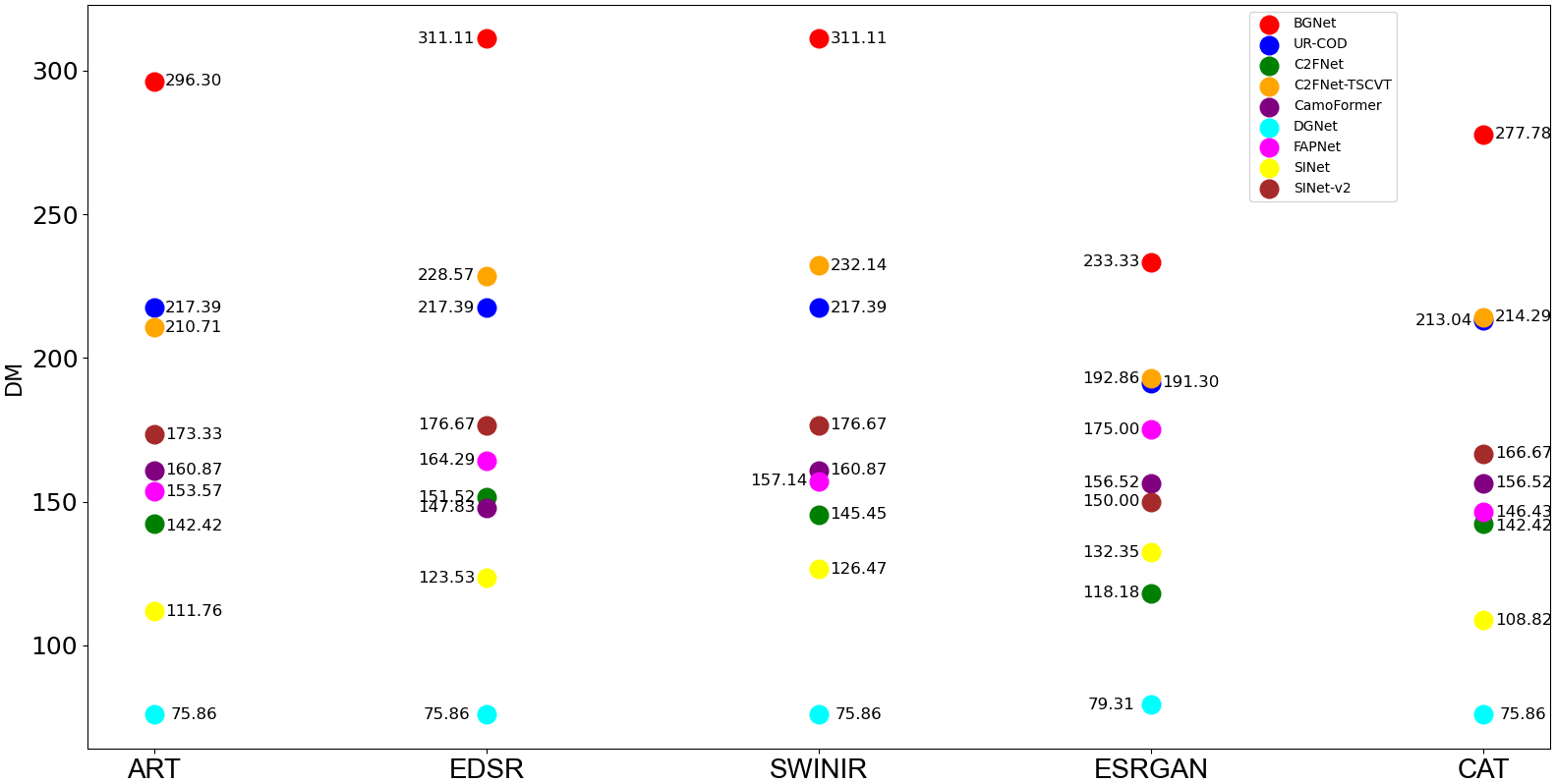} 
  \caption{scatter plot of DM on CHAMELEON}
\end{figure}
Our observations are as follows:\\
\begin{itemize}[label=$\bullet$, align=left, leftmargin=*]
\item Overall, considering the input data that undergoes downsampling followed by super-resolution, DGNet \cite{ji2023deep} exhibits the least amount of performance degradation. This may be attributed to DGNet's \cite{ji2023deep}ability to detect texture patterns while suppressing the noise from irrelevant background by an intensity-sensitive strategy focusing on the gradient inside the camouflaged object. More specifically, DGnet \cite{ji2023deep} consists of two connected learning branches, i.e., context encoder and texture encoder. The former extracts high-level features from images by picking out the top-three features from the widely used EfficientNet \cite{tan2019efficientnet}. The latter is a structural texture extractor supervised by the object-level gradient map. This map holds both the gradient cues of the object’s boundaries and interior regions while neglecting irrelevant background noises. When image quality deteriorates, the increase in irrelevant environmental noise can significantly impact the performance of COD models. We speculate that DGNet's \cite{ji2023deep} design for suppressing background noise allows it to outperform other COD models in this experiment. Its robustness makes it a promising architecture for further exploration of its potential abilities. \\
\item We found that BGNet \cite{sun2022boundary} struggled in the experiment, while UR-COD \cite{kajiura2021improving}, which also emphasizes edge
detection, exhibited better performance than BGNet \cite{sun2022boundary}. Based on our analysis, this is likely because the UR-COD model \cite{kajiura2021improving} uses an uncertainty-aware map refinement module (UAMR) to absorb uncertainty. More specifically, the UR-COD model \cite{kajiura2021improving} firstly takes the input image to both a pseudo-edge generator and a conventional COD model (in our work, we use SINet-v2 \cite{fan2021concealed}). The former explicitly estimates the boundaries of camouflaged objects and produces a pseudo-edge label, while the latter generates a camouflaged object detection result, which is treated as a pseudo-map label. However, these pseudo labels are generated by learning-based models, thus containing uncertainty. So UR-COD \cite{kajiura2021improving} extended UCNet \cite{zhang2020uc}and designed a UAMR module to reduce uncertainty and noise. In this experiment, in addition to the uncertainty introduced by pseudo labels, the input image itself contains noise and uncertainty after being processed through downsampling and SR models. So we speculate that this module that mitigates noise and uncertainty played a pivotal role in the superior performance of UR-COD \cite{kajiura2021improving} compared to BGNet \cite{sun2022boundary}.In future research, if the input images contain noise, we believe that the approach employed by UR-COD \cite{kajiura2021improving} is worth considering and adopting as a reference. \\
\item Despite its less notable performance on the DM metric, CamoFormer \cite{yin2022camoformer} exhibits commendable results on the Avg\_D metric, ranking just below DGNet \cite{ji2023deep}. Since CamoFormer \cite{yin2022camoformer} is a transformer-based model, this result could potentially be attributed to the long-range modelling capabilities of the Transformer architecture. Furthermore, it is noteworthy that among these nine models, CamoFormer \cite{yin2022camoformer} demonstrates the best performance when other conditions are held constant. This implies that CamoFormer \cite{yin2022camoformer} exhibits commendable robustness while maintaining high performance. Therefore, we believe that the Transformer architecture will remain a popular research direction in the COD field for the foreseeable future.
\end{itemize}
\subsection{Summary}
From Figure 1 to Figure 8, it can be observed that the ART method achieved the best results in terms of Avg\_D among the four COD datasets processed using five SR methods and subsequently by nine COD models. In terms of DM, the CAT method yielded the best results. These findings are consistent with the ranking of the performance indicators for the five SR methods across the four COD datasets. Notably, both of the top-performing SR methods employed transformer models, as did the CamoFormer model, which demonstrated the best performance among the nine COD models. These results suggest that the transformer model has become the de facto choice in natural language processing, while also exhibiting strong generalization and robustness in the fields of SR and COD. Therefore, the transformer model is likely to remain a popular research direction in the domains of SR and COD for the next few years.






\section{Discussion and Prospects}
\begin{itemize}[label=$\bullet$, align=left, leftmargin=*]
\item Disguised targets often blur the edges and details, making them harder to detect in low-resolution images. By utilizing SR processing, we can partially restore the edge and detail information of the target, providing clearer visual cues that can help detection models more accurately locate and identify the disguised targets. 
\item The transformer model exhibits strong generalization and robustness in both the SR and COD domains. This suggests that the transformer model is likely to remain a popular research direction in the SR and COD fields in the coming years.\\
\textbf{Our work has two main limitations:} 
\item \textbf{Firstly}, the selection of COD and SR models may not be comprehensive enough, as the experiments conducted did not encompass all possible models. \textbf{Secondly}, the metric proposed for assessing the degradation in performance of the COD model may not be optimal or scientifically rigorous.
\item \textbf{In the future}, our plan entails exploring and developing superior SR and COD models by means of evaluating performance indicators. Additionally, we aim to investigate and refine assessment metrics to ensure a more comprehensive and accurate evaluation of model performance.
\end{itemize}
\section{Conclusion}
In this paper, \textbf{we present a novel comparative study on SR and COD}. \textbf{Initially}, we analyze and compare the performance differences of various SR models on the COD datasets. \textbf{Subsequently}, we quantitatively measure the degree of performance degradation of different COD models on datasets processed by different SR methods. This provides a new perspective for evaluating SR methods and COD performance, respectively. \textbf{Finally}, we summarize the observed patterns in the experiments and discuss the limitations of our work. We hope that this article will provide new ideas and insights for researchers in both communities in the future.

\small{\bibliographystyle{plainnat}
\bibliography{references}}
\begin{enumerate}
\newpage
\section*{Checklist}


\item For all authors...
\begin{enumerate}
  \item Do the main claims made in the abstract and introduction accurately reflect the paper's contributions and scope?
    \answerYes{}
  \item Did you describe the limitations of your work?
    \answerYes{}
  \item Did you discuss any potential negative societal impacts of your work?
  \answerNo{There are no potential negative societal impacts in this paper because we used synthetic graph generators. }
  \item Have you read the ethics review guidelines and ensured that your paper conforms to them?
    \answerYes{}
\end{enumerate}

\item If you are including theoretical results...
\begin{enumerate}
  \item Did you state the full set of assumptions of all theoretical results?
    \answerNA{}
	\item Did you include complete proofs of all theoretical results?
    \answerNA{}
\end{enumerate}

\item If you ran experiments (e.g. for benchmarks)...
\begin{enumerate}
  \item Did you include the code, data, and instructions needed to reproduce the main experimental results (either in the supplemental material or as a URL)?
    \answerYes{}
  \item Did you specify all the training details (e.g., data splits, hyperparameters, how they were chosen)?
    \answerYes{}
	\item Did you report error bars (e.g., with respect to the random seed after running experiments multiple times)?
    \answerYes{}
	\item Did you include the total amount of compute and the type of resources used (e.g., type of GPUs, internal cluster, or cloud provider)?
    \answerYes{}
\end{enumerate}

\item If you are using existing assets (e.g., code, data, models) or curating/releasing new assets...
\begin{enumerate}
  \item If your work uses existing assets, did you cite the creators?
    \answerYes{}
  \item Did you mention the license of the assets?
    \answerYes{}
  \item Did you include any new assets either in the supplemental material or as a URL?
    \answerYes{}
  \item Did you discuss whether and how consent was obtained from people whose data you're using/curating?
    \answerNA{}
  \item Did you discuss whether the data you are using/curating contains personally identifiable information or offensive content?
    \answerNA{}
\end{enumerate}

\item If you used crowdsourcing or conducted research with human subjects...
\begin{enumerate}
  \item Did you include the full text of instructions given to participants and screenshots, if applicable?
    \answerNA{}
  \item Did you describe any potential participant risks, with links to Institutional Review Board (IRB) approvals, if applicable?
    \answerNA{}
  \item Did you include the estimated hourly wage paid to participants and the total amount spent on participant compensation?
    \answerNA{}
\end{enumerate}

\end{enumerate}


\newpage
\section{Appendix}

Include extra information in the appendix. This section will often be part of the supplemental material. Please see the call on the NeurIPS website for links to additional guides on dataset publication.

\begin{enumerate}
   \item Submission introducing new datasets must include the following in the supplementary materials:
        \begin{enumerate}
        \item Dataset documentation and intended uses. Recommended documentation frameworks include datasheets for datasets, dataset nutrition labels, data statements for NLP, and accountability frameworks.
        \item URL to website/platform where the dataset/benchmark can be viewed and downloaded by the reviewers.
        \item Author statement that they bear all responsibility in case of violation of rights, etc., and confirmation of the data license.
        \item Hosting, licensing, and maintenance plan. The choice of hosting platform is yours, as long as you ensure access to the data (possibly through a curated interface) and will provide the necessary maintenance.
        \end{enumerate}

    \item To ensure accessibility, the supplementary materials for datasets must include the following:
    \begin{enumerate}
    \item Links to access the dataset and its metadata. This can be hidden upon submission if the dataset is not yet publicly available but must be added in the camera-ready version. In select cases, \eg when the data can only be released at a later date, this can be added afterward. Simulation environments should link to (open source) code repositories.
    \item The dataset itself should ideally use an open and widely used data format. Provide a detailed explanation on how the dataset can be read. For simulation environments, use existing frameworks or explain how they can be used.
    \item Long-term preservation: It must be clear that the dataset will be available for a long time, either by uploading to a data repository or by explaining how the authors themselves will ensure this.
    \item Explicit license: Authors must choose a license, ideally a CC license for datasets, or an open source license for code (\eg. RL environments).
    \item Add structured metadata to a dataset's meta-data page using Web standards (like schema.org and DCAT): This allows it to be discovered and organized by anyone. If you use an existing data repository, this is often done automatically.
    \item Highly recommended: a persistent dereferenceable identifier (\eg. a DOI minted by a data repository or a prefix on identifiers.org) for datasets, or a code repository (\eg. GitHub, GitLab,...) for code. If this is not possible or useful, please explain why.
    \end{enumerate}

    \item For benchmarks, the supplementary materials must ensure that all results are easily reproducible. Where possible, use a reproducibility framework such as the ML reproducibility checklist, or otherwise guarantee that all results can be easily reproduced, i.e. all necessary datasets, code, and evaluation procedures must be accessible and documented.

    \item For papers introducing best practices in creating or curating datasets and benchmarks, the above supplementary materials are not required.
\end{enumerate}

 \begin{table}
    \caption{\textbf{Quantitative comparison when datasets are not preprocessed by any SR method.} The symbols $\uparrow$/$\downarrow$ indicate that a higher/lower score is better.}
    \centering
    \resizebox{\textwidth}{!}{
    \begin{tabular}{*{17}{c}}
        \toprule
        \multirow{4}{*}{Method} & \multicolumn{4}{c}{CAMO} & \multicolumn{4}{c}{CHAMELEON} & \multicolumn{4}{c}{COD10K} & \multicolumn{4}{c}{NC4K} \\
        \cmidrule(lr){2-5} \cmidrule(lr){6-9} \cmidrule(lr){10-13} \cmidrule(lr){14-17}
        & $\textit{S}_{\alpha}\uparrow$ & $\textit{E}_{\phi}\uparrow$ & $\textit{F}^{w}_{\beta}\uparrow$ & $\textit{M}\downarrow$ & $\textit{S}_{\alpha}\uparrow$ & $\textit{E}_{\phi}\uparrow$ & $\textit{F}^{w}_{\beta}\uparrow$ & $\textit{M}\downarrow$ & $\textit{S}_{\alpha}\uparrow$ & $\textit{E}_{\phi}\uparrow$ & $\textit{F}^{w}_{\beta}\uparrow$ & $\textit{M}\downarrow$ & $\textit{S}_{\alpha}\uparrow$ & $\textit{E}_{\phi}\uparrow$ & $\textit{F}^{w}_{\beta}\uparrow$ & $\textit{M}\downarrow$ \\
        \midrule
        BGnet & 0.812 & 0.870 & 0.749 & 0.073 & 0.901 & 0.943 & 0.850 & 0.027 & 0.831 & 0.901 & 0.722 & 0.033 & 0.851 & 0.907 & 0.788 & 0.044 \\
        UR-COD & 0.814 & 0.891 & 0.758 & 0.067 & 0.901 & 0.960 & 0.862 & 0.023 & 0.816 & 0.903 & 0.708 & 0.033 & 0.844 & 0.910 & 0.787 & 0.045 \\
        C2FNet & 0.796 & 0.854 & 0.719 & 0.080 & 0.886 & 0.933 & 0.825 & 0.033 & 0.813 & 0.890 & 0.686 & 0.036 & 0.838 & 0.897 & 0.762 & 0.049 \\
        C2FNet-TSCVT & 0.799 & 0.859 & 0.730 & 0.077 & 0.893 & 0.947 & 0.845 & 0.028 & 0.811 & 0.887 & 0.691 & 0.036 & 0.840 & 0.896 & 0.770 & 0.048 \\
        CamoFormer & 0.872 & 0.929 & 0.831 & 0.046 & 0.910 & 0.956 & 0.859 & 0.023 & 0.869 & 0.932 & 0.786 & 0.023 & 0.892 & 0.939 & 0.847 & 0.030 \\
        DGnet & 0.839 & 0.901 & 0.769 & 0.057 & 0.890 & 0.938 & 0.816 & 0.029 & 0.822 & 0.896 & 0.693 & 0.033 & 0.857 & 0.911 & 0.784 & 0.042 \\
        FAPNet & 0.815 & 0.865 & 0.734 & 0.076 & 0.893 & 0.940 & 0.825 & 0.028 & 0.822 & 0.888 & 0.694 & 0.036 & 0.851 & 0.899 & 0.775 & 0.047 \\
        SINet & 0.745 & 0.804 & 0.644 & 0.092 & 0.872 & 0.936 & 0.806 & 0.034 & 0.776 & 0.864 & 0.631 & 0.043 & 0.808 & 0.871 & 0.723 & 0.058 \\
        SINet-v2 & 0.820 & 0.882 & 0.743 & 0.070 & 0.888 & 0.941 & 0.816 & 0.030 & 0.815 & 0.887 & 0.680 & 0.037 & 0.847 & 0.903 & 0.770 & 0.048 \\
        \bottomrule
    \end{tabular}
    }
\end{table}

\begin{table}
    \caption{\textbf{Quantitative comparison when datasets are preprocessed by ART method.} The symbols $\uparrow$/$\downarrow$ indicate that a higher/lower score is better.}
    \centering
    \resizebox{\textwidth}{!}{
    \begin{tabular}{*{17}{c}}
       \toprule
       \multirow{2}*{Method} & \multicolumn{4}{c}{CAMO} & \multicolumn{4}{c}{CHAMELEON}  &\multicolumn{4}{c}{COD10K} &\multicolumn{4}{c}{NC4K} \\
       \cmidrule (lr){2-5} \cmidrule (lr){6-9} \cmidrule (lr){10-13} \cmidrule (lr){14-17}
       & $\textit{S}_{\alpha}\uparrow$ & $\textit{E}_{\phi}\uparrow$ & $\textit{F}^{w}_{\beta}\uparrow$ & $\textit{M}\downarrow$ & $\textit{S}_{\alpha}\uparrow$ & $\textit{E}_{\phi}\uparrow$ & $\textit{F}^{w}_{\beta}\uparrow$ & $\textit{M}\downarrow$ & $\textit{S}_{\alpha}\uparrow$ & $\textit{E}_{\phi}\uparrow$ & $\textit{F}^{w}_{\beta}\uparrow$ & $\textit{M}\downarrow$ & $\textit{S}_{\alpha}\uparrow$ & $\textit{E}_{\phi}\uparrow$ & $\textit{F}^{w}_{\beta}\uparrow$ & $\textit{M}\downarrow$ \\
       \midrule
       BGnet & 0.655 & 0.734 & 0.499 & 0.152  & 0.717  & 0.783  & 0.559  & 0.107  & 0.705  & 0.787 & 0.507 & 0.073  & 0.715  & 0.786  & 0.567  & 0.106   \\
       UR-COD & 0.693  & 0.777   & 0.573 & 0.131  & 0.778  & 0.855   & 0.665   & 0.073  & 0.723  & 0.811  & 0.547   & 0.065   & 0.741   & 0.818  & 0.619   & 0.090 \\    
       C2FNet & 0.660   & 0.717  & 0.503  & 0.130   & 0.748  & 0.817   & 0.605 & 0.080   & 0.701  & 0.782  & 0.500 & 0.069   & 0.726  & 0.796   & 0.586  & 0.092 \\
       C2FNet-TSCVT & 0.652  & 0.720   & 0.492  & 0.136   & 0.733  & 0.805   & 0.589   & 0.087  & 0.696  & 0.774  & 0.496  & 0.076   & 0.719  & 0.789   & 0.577   & 0.098 \\
       CamoFormer & 0.766  & 0.835   & 0.661  & 0.094  & 0.806   & 0.867  & 0.693   & 0.060    & 0.790  & 0.862   & 0.652  & 0.046  & 0.804  & 0.864   & 0.706   & 0.065 \\
       DGnet & 0.738  & 0.799 & 0.607  & 0.101 & 0.821  & 0.891 & 0.697  & 0.051  & 0.763  & 0.839  & 0.589 & 0.050  & 0.786 & 0.850  & 0.667  & 0.069 \\
       FAPNet & 0.704  & 0.761   & 0.551  & 0.130   & 0.793  & 0.847   & 0.644  & 0.071  & 0.732  & 0.789  & 0.527  & 0.070  & 0.753  & 0.805  & 0.603  & 0.092 \\
       SINet & 0.640  & 0.682  & 0.463  & 0.130  & 0.752  & 0.816  & 0.612  & 0.072  & 0.708  & 0.796  & 0.513 & 0.061 & 0.724  & 0.794  & 0.586  & 0.088 \\
       SINet-v2 & 0.702  & 0.761  & 0.558  & 0.140  & 0.778  & 0.833  & 0.634  & 0.082  & 0.725  & 0.791  & 0.524  & 0.072  & 0.746  & 0.805  & 0.601  & 0.096\\ 
       \bottomrule
    \end{tabular}
    }
\end{table}

\begin{table}
    \caption{\textbf{Quantitative comparison when datasets are preprocessed by EDSR method.} The symbols $\uparrow$/$\downarrow$ indicate that a higher/lower score is better.}
    \centering
    \resizebox{\textwidth}{!}{
    \begin{tabular}{*{17}{c}}
        \toprule
        \multirow{2}*{Method} & \multicolumn{4}{c}{CAMO} & \multicolumn{4}{c}{CHAMELEON}  & \multicolumn{4}{c}{COD10K} & \multicolumn{4}{c}{NC4K} \\
        \cmidrule(lr){2-5} \cmidrule(lr){6-9} \cmidrule(lr){10-13} \cmidrule(lr){14-17}
        & $\textit{S}_{\alpha}\uparrow$ & $\textit{E}_{\phi}\uparrow$ & $\textit{F}^{w}_{\beta}\uparrow$ & $\textit{M}\downarrow$ & $\textit{S}_{\alpha}\uparrow$ & $\textit{E}_{\phi}\uparrow$ & $\textit{F}^{w}_{\beta}\uparrow$ & $\textit{M}\downarrow$ & $\textit{S}_{\alpha}\uparrow$ & $\textit{E}_{\phi}\uparrow$ & $\textit{F}^{w}_{\beta}\uparrow$ & $\textit{M}\downarrow$ & $\textit{S}_{\alpha}\uparrow$ & $\textit{E}_{\phi}\uparrow$ & $\textit{F}^{w}_{\beta}\uparrow$ & $\textit{M}\downarrow$ \\
        \midrule
        BGnet & 0.653 & 0.735 & 0.493 & 0.152 & 0.705 & 0.772 & 0.542 & 0.111 & 0.698 & 0.778 & 0.496 & 0.077 & 0.708 & 0.778 & 0.556 & 0.110 \\
        UR-COD & 0.689 & 0.773 & 0.569 & 0.136 & 0.779 & 0.852 & 0.663 & 0.073 & 0.722 & 0.808 & 0.544 & 0.066 & 0.739 & 0.815 & 0.615 & 0.092 \\
        C2FNet & 0.648 & 0.709 & 0.482 & 0.135 & 0.745 & 0.811 & 0.599 & 0.083 & 0.701 & 0.778 & 0.497 & 0.070 & 0.725 & 0.793 & 0.582 & 0.093 \\
        C2FNet-TSCVT & 0.643 & 0.707 & 0.478 & 0.140 & 0.722 & 0.790 & 0.568 & 0.092 & 0.693 & 0.769 & 0.491 & 0.077 & 0.715 & 0.785 & 0.570 & 0.099 \\
        CamoFormer & 0.760 & 0.830 & 0.655 & 0.099 & 0.809 & 0.875 & 0.697 & 0.057 & 0.788 & 0.859 & 0.649 & 0.047 & 0.803 & 0.862 & 0.704 & 0.066 \\
        DGnet & 0.738 & 0.801 & 0.608 & 0.103 & 0.822 & 0.886 & 0.698 & 0.051 & 0.762 & 0.839 & 0.588 & 0.050 & 0.785 & 0.849 & 0.665 & 0.070 \\
        FAPNet & 0.706 & 0.761 & 0.551 & 0.132 & 0.793 & 0.842 & 0.640 & 0.074 & 0.731 & 0.787 & 0.525 & 0.071 & 0.750 & 0.802 & 0.597 & 0.093 \\
        SINet & 0.636 & 0.682 & 0.458 & 0.133 & 0.747 & 0.810 & 0.604 & 0.076 & 0.708 & 0.794 & 0.513 & 0.062 & 0.724 & 0.794 & 0.586 & 0.088 \\
        SINet-v2 & 0.698 & 0.756 & 0.552 & 0.146 & 0.778 & 0.831 & 0.632 & 0.083 & 0.723 & 0.789 & 0.522 & 0.072 & 0.744 & 0.801 & 0.597 & 0.099\\
        \bottomrule
    \end{tabular}
    }
\end{table}

\begin{table}
    \caption{\textbf{Quantitative comparison when datasets are preprocessed by SWINIR method.} The symbols $\uparrow$/$\downarrow$ indicate that a higher/lower score is better.}
    \centering
    \resizebox{\textwidth}{!}{
    \begin{tabular}{*{17}{c}}
       \toprule
       \multirow{2}*{Method} & \multicolumn{4}{c}{CAMO} & \multicolumn{4}{c}{CHAMELEON}  &\multicolumn{4}{c}{COD10K} &\multicolumn{4}{c}{NC4K} \\
       \cmidrule (lr){2-5} \cmidrule (lr){6-9} \cmidrule (lr){10-13} \cmidrule (lr){14-17}
       & $\textit{S}_{\alpha}\uparrow$ & $\textit{E}_{\phi}\uparrow$ & $\textit{F}^{w}_{\beta}\uparrow$ & $\textit{M}\downarrow$ & $\textit{S}_{\alpha}\uparrow$ & $\textit{E}_{\phi}\uparrow$ & $\textit{F}^{w}_{\beta}\uparrow$ & $\textit{M}\downarrow$ & $\textit{S}_{\alpha}\uparrow$ & $\textit{E}_{\phi}\uparrow$ & $\textit{F}^{w}_{\beta}\uparrow$ & $\textit{M}\downarrow$ & $\textit{S}_{\alpha}\uparrow$ & $\textit{E}_{\phi}\uparrow$ & $\textit{F}^{w}_{\beta}\uparrow$ & $\textit{M}\downarrow$ \\
       \midrule
       BGnet & 0.651 & 0.733 & 0.492 & 0.152  & 0.709  & 0.774  & 0.546  & 0.111  & 0.698  & 0.779 & 0.497 & 0.076  & 0.710  & 0.781  & 0.558  & 0.108   \\
       UR-COD & 0.690  & 0.778   & 0.568 & 0.134  & 0.776  & 0.853   & 0.658   & 0.073  & 0.722  & 0.809  & 0.545 & 0.066  & 0.738   & 0.815   & 0.615  & 0.091 \\    
       C2FNet & 0.652   & 0.709  & 0.491  & 0.133   & 0.745  & 0.807   & 0.597 & 0.081   & 0.700  & 0.778  & 0.496 & 0.071   & 0.722  & 0.792   & 0.579  & 0.094 \\
       C2FNet-TSCVT & 0.651  & 0.710   & 0.491  & 0.141   & 0.720  & 0.790   & 0.569   & 0.093  & 0.694  & 0.773  & 0.493  & 0.077   & 0.715  & 0.785   & 0.570   & 0.100 \\
       CamoFormer & 0.761  & 0.832   & 0.657  & 0.098  & 0.806   & 0.867  & 0.693   & 0.060    & 0.788  & 0.861   & 0.648  & 0.046  & 0.802  & 0.861   & 0.702   & 0.067 \\
       DGnet & 0.736  & 0.799 & 0.604  & 0.103 & 0.821  & 0.887 & 0.696  & 0.051  & 0.761  & 0.837  & 0.586 & 0.050  & 0.784 & 0.849  & 0.664  & 0.070 \\
       FAPNet & 0.706  & 0.765   & 0.553  & 0.130   & 0.793  & 0.836   & 0.638  & 0.072  & 0.730  & 0.785  & 0.523  & 0.072  & 0.749  & 0.801  & 0.596  & 0.093 \\
       SINet & 0.638  & 0.684  & 0.461  & 0.131  & 0.746  & 0.815  & 0.599  & 0.077  & 0.708  & 0.795  & 0.514 & 0.062 & 0.724  & 0.795  & 0.586  & 0.089 \\
       SINet-v2 & 0.698  & 0.760  & 0.552  & 0.144  & 0.775  & 0.831  & 0.627  & 0.083  & 0.723  & 0.789  & 0.522  & 0.072  & 0.744  & 0.802  & 0.597  & 0.098\\ 
       \bottomrule
    \end{tabular}
    }
\end{table}

\begin{table}
    \caption{\textbf{Quantitative comparison when datasets are preprocessed by ESRGAN method.} The symbols $\uparrow$/$\downarrow$ indicate that a higher/lower score is better.}
    \centering
    \resizebox{\textwidth}{!}
    {
    \begin{tabular}{*{17}{c}}
       \toprule
       \multirow{2}*{Method} & \multicolumn{4}{c}{CAMO} & \multicolumn{4}{c}{CHAMELEON}  &\multicolumn{4}{c}{COD10K} &\multicolumn{4}{c}{NC4K} \\
       \cmidrule (lr){2-5} \cmidrule (lr){6-9} \cmidrule (lr){10-13} \cmidrule (lr){14-17}
       & $\textit{S}_{\alpha}\uparrow$ & $\textit{E}_{\phi}\uparrow$ & $\textit{F}^{w}_{\beta}\uparrow$ & $\textit{M}\downarrow$ & $\textit{S}_{\alpha}\uparrow$ & $\textit{E}_{\phi}\uparrow$ & $\textit{F}^{w}_{\beta}\uparrow$ & $\textit{M}\downarrow$ & $\textit{S}_{\alpha}\uparrow$ & $\textit{E}_{\phi}\uparrow$ & $\textit{F}^{w}_{\beta}\uparrow$ & $\textit{M}\downarrow$ & $\textit{S}_{\alpha}\uparrow$ & $\textit{E}_{\phi}\uparrow$ & $\textit{F}^{w}_{\beta}\uparrow$ & $\textit{M}\downarrow$ \\
       \midrule
       BGnet & 0.634 & 0.708 & 0.464 & 0.143  & 0.713  & 0.790  & 0.556  & 0.090  & 0.689  & 0.780 & 0.483 & 0.073  & 0.694  & 0.776  & 0.538  & 0.100   \\
       UR-COD & 0.660  & 0.763   & 0.528 & 0.138  & 0.780  & 0.865   & 0.660   & 0.067  & 0.718  & 0.814  & 0.537   & 0.063   & 0.721   & 0.805  & 0.589   & 0.093 \\    
       C2FNet & 0.635   & 0.684  & 0.461  & 0.135   & 0.754  & 0.831   & 0.619  & 0.072   & 0.700  & 0.783  & 0.495 & 0.066   & 0.711  & 0.784   & 0.563  & 0.093 \\
       C2FNet-TSCVT & 0.626  & 0.692   & 0.456  & 0.142   & 0.733  & 0.814   & 0.590   & 0.082  & 0.689  & 0.774  & 0.485  & 0.072   & 0.696  & 0.773   & 0.543   & 0.099 \\
       CamoFormer & 0.725  & 0.804   & 0.607  & 0.107  & 0.798   & 0.877  & 0.685   & 0.059    & 0.779  & 0.857   & 0.634  & 0.046  & 0.785  & 0.850   & 0.677   & 0.070 \\
       DGnet & 0.725  & 0.791 & 0.591  & 0.109 & 0.818  & 0.877 & 0.695  & 0.052  & 0.758  & 0.837  & 0.581 & 0.051  & 0.778 & 0.846  & 0.655  & 0.072 \\
       FAPNet & 0.679  & 0.733   & 0.509  & 0.136   & 0.783  & 0.836   & 0.628  & 0.077  & 0.723  & 0.785  & 0.513  & 0.072  & 0.732  & 0.789  & 0.572  & 0.097 \\
       SINet & 0.622  & 0.670  & 0.437  & 0.134  & 0.728  & 0.795  & 0.577  & 0.079  & 0.700  & 0.787  & 0.500 & 0.063 & 0.710  & 0.784  & 0.567  & 0.090 \\
       SINet-v2 & 0.672  & 0.751  & 0.518  & 0.146  & 0.780  & 0.842  & 0.630 & 0.075  & 0.720  & 0.792  & 0.514  & 0.070  & 0.729  & 0.793  & 0.573  & 0.099\\ 
       \bottomrule
     \end{tabular}
    }
\end{table}

\begin{table}
    \caption{\textbf{Quantitative comparison when datasets are preprocessed by CAT method.} The symbols $\uparrow$/$\downarrow$ indicate that a higher/lower score is better.}
    \centering
    \resizebox{\textwidth}{!}
    {
    \begin{tabular}{*{17}{c}}
       \toprule
       \multirow{2}*{Method} & \multicolumn{4}{c}{CAMO} & \multicolumn{4}{c}{CHAMELEON}  &\multicolumn{4}{c}{COD10K} &\multicolumn{4}{c}{NC4K} \\
       \cmidrule (lr){2-5} \cmidrule (lr){6-9} \cmidrule (lr){10-13} \cmidrule (lr){14-17}
       & $\textit{S}_{\alpha}\uparrow$ & $\textit{E}_{\phi}\uparrow$ & $\textit{F}^{w}_{\beta}\uparrow$ & $\textit{M}\downarrow$ & $\textit{S}_{\alpha}\uparrow$ & $\textit{E}_{\phi}\uparrow$ & $\textit{F}^{w}_{\beta}\uparrow$ & $\textit{M}\downarrow$ & $\textit{S}_{\alpha}\uparrow$ & $\textit{E}_{\phi}\uparrow$ & $\textit{F}^{w}_{\beta}\uparrow$ & $\textit{M}\downarrow$ & $\textit{S}_{\alpha}\uparrow$ & $\textit{E}_{\phi}\uparrow$ & $\textit{F}^{w}_{\beta}\uparrow$ & $\textit{M}\downarrow$ \\
       \midrule
       BGnet & 0.656 & 0.732 & 0.500 & 0.152  & 0.719  & 0.785  & 0.564  & 0.102  & 0.704  & 0.787 & 0.506 & 0.073  & 0.715  & 0.785  & 0.567  & 0.107   \\
       UR-COD & 0.693  & 0.780   & 0.573 & 0.131  & 0.783  & 0.856   & 0.672   & 0.072  & 0.723  & 0.810  & 0.547   & 0.065   & 0.742   & 0.819  & 0.622   & 0.089 \\    
       C2FNet & 0.656   & 0.712  & 0.498  & 0.132   & 0.750  & 0.817   & 0.602  & 0.080   & 0.701  & 0.782  & 0.499 & 0.069   & 0.727  & 0.796   & 0.586  & 0.092 \\
       C2FNet-TSCVT & 0.650  & 0.719   & 0.489  & 0.138   & 0.729  & 0.796   & 0.583   & 0.088  & 0.697  & 0.775  & 0.497  & 0.075   & 0.719  & 0.791   & 0.578   & 0.098 \\
       CamoFormer & 0.764  & 0.833   & 0.660  & 0.096  & 0.809   & 0.870  & 0.696   & 0.059    & 0.791  & 0.863   & 0.653  & 0.045  & 0.805  & 0.865   & 0.708   & 0.065 \\
       DGnet & 0.736  & 0.797 & 0.605  & 0.102 & 0.822  & 0.889 & 0.698  & 0.051  & 0.763  & 0.839  & 0.588 & 0.050  & 0.786 & 0.850  & 0.667  & 0.069 \\
       FAPNet & 0.711  & 0.771   & 0.561  & 0.127   & 0.800  & 0.850   & 0.656  & 0.069  & 0.732  & 0.789  & 0.528  & 0.070  & 0.754  & 0.806  & 0.605  & 0.091 \\
       SINet & 0.643  & 0.696  & 0.471  & 0.129  & 0.756  & 0.816  & 0.621  & 0.071  & 0.708  & 0.796  & 0.514 & 0.061 & 0.726  & 0.796  & 0.590  & 0.087 \\
       SINet-v2 & 0.702  & 0.764  & 0.558  & 0.141  & 0.781  & 0.835  & 0.641  & 0.080  & 0.725  & 0.791  & 0.525  & 0.072  & 0.748  & 0.806  & 0.604  & 0.096\\ 
       \bottomrule
     \end{tabular}
    }
\end{table}

\begin{table}
\center
\caption{COD-SR-BGNet}
\renewcommand{\arraystretch}{1.5}
\resizebox{150mm}{55mm}{ 
\begin{tabular}{|c|c|c|c|c|c|c|c|c|c|c|c|} \hline

\multirow{5}{*}{origin} &{Dataset} &{method} &{Smeasure} &{wFmeasure} &{MAE}  &{adpEm}  &{meanEm} &{maxEm} &{adpFm}    &{meanFm} &{maxFm} \\  \cline{2-12}
& {CAMO} & {BGNet} & {0.812} & {0.749}  & 0.073 & 0.876 & 0.870 & 0.882  & 0.786 & 0.789 & 0.799     \\  \cline{2-12}
& {CHAMELEON} & {BGNet} & 0.901 & 0.85	& 0.027 	& 0.938 & 0.943 & 0.954  & 0.846 & 0.86  & 0.882  \\ \cline{2-12}
& {COD10K} & {BGNet} & 0.831 & 0.722	& 0.033 	& 0.902 & 0.901 & 0.911  & 0.739 & 0.753  & 0.774  \\ \cline{2-12}
& {NC4K} & {BGNet} & 0.851 & 0.788	& 0.044 	& 0.911 & 0.907 & 0.916  & 0.813 & 0.820  & 0.833  \\ \hline
\multirow{4}{*}{ART} 
& {CAMO} & {BGNet} & {0.655} & {0.499}  & 0.152 & 0.755 & 0.734 & 0.746  & 0.567 & 0.56 & 0.567     \\  \cline{2-12}
& {CHAMELEON} & {BGNet} & 0.717 & 0.559	& 0.107 	& 0.803 & 0.783 & 0.805  & 0.612 & 0.605  & 0.616  \\ \cline{2-12}
& {COD10K} & {BGNet} & 0.705 & 0.507	& 0.073 	& 0.786 & 0.787 & 0.801  & 0.548 & 0.554  & 0.574  \\ \cline{2-12}
& {NC4K} & {BGNet} & 0.715 & 0.567	& 0.106 	& 0.793 & 0.786 & 0.798  & 0.617 & 0.619  & 0.638  \\ \hline

\multirow{4}{*}{EDSR} &{CAMO} &{BGNet} &0.653 &0.493  &0.152  &0.755  &0.735  &0.747  &0.559  &0.551 &0.558 \\ \cline{2-12}

&{CHAMELEON} &{BGNet} &0.705 &0.542	&0.111 	&0.788 &0.772 &0.786  &0.578 &0.586  &0.597  \\ \cline{2-12}
&{COD10K} &{BGNet} &0.698 &0.496	&0.077 	&0.777 &0.778 &0.792  &0.536 &0.542  &0.563  \\ \cline{2-12}
&{NC4K} &{BGNet} &0.708 &0.556	&0.11 	&0.786 &0.778 &0.792  &0.607 &0.607  &0.627  \\ \hline

\multirow{4}{*}{SWINIR} &{CAMO} &{BGNet} &0.651 &0.492  &0.152  &0.758  &0.733  &0.745  &0.564  &0.549 &0.555 \\ \cline{2-12}

&{CHAMELEON} &{BGNet} &0.709 &0.546	&0.111 	&0.794 &0.774 &0.794  &0.594 &0.594  &0.603  \\ \cline{2-12}
&{COD10K} &{BGNet} &0.698 &0.497	&0.076 	&0.779 &0.779 &0.794  &0.538 &0.544  &0.566  \\ \cline{2-12}
&{NC4K} &{BGNet} &0.71 &0.558	&0.108 	&0.788 &0.781 &0.793  &0.61 &0.61  &0.627  \\ \hline

\multirow{4}{*}{ESRGAN}&{CAMO} &{BGNet} &0.634 &0.464  &0.143  &0.735  &0.708  &0.722  &0.546  &0.529 &0.536 \\ \cline{2-12}

&{CHAMELEON} &{BGNet} &0.713 &0.556	&0.09 	&0.806 &0.79 &0.802  &0.621 &0.612  &0.619  \\ \cline{2-12}
&{COD10K} &{BGNet} &0.689 &0.483	&0.073 	&0.784 &0.78 &0.789  &0.53 &0.534  &0.55  \\ \cline{2-12}
&{NC4K} &{BGNet} &0.694 &0.538	&0.1 	&0.792 &0.776 &0.785  &0.603 &0.597  &0.604  \\ \hline

\multirow{4}{*}{CAT}&{CAMO} &{BGNet} &0.656 &0.5  &0.152  &0.751  &0.732  &0.743  &0.563  &0.558 &0.566 \\ \cline{2-12}

&{CHAMELEON} &{BGNet} &0.719 &0.564	&0.102 	&0.804 &0.785 &0.802  &0.605 &0.609  &0.62  \\ \cline{2-12}
&{COD10K} &{BGNet} &0.704 &0.506	&0.073 	&0.785 &0.787 &0.8  &0.547 &0.553  &0.574  \\ \cline{2-12}
&{NC4K} &{BGNet} &0.715 &0.567	&0.107 	&0.792 &0.785 &0.799  &0.617 &0.618  &0.638  \\ \hline
\end{tabular}
}	
\end{table}

\begin{table}
\center
\caption{COD-SR-UR-SINetv2}
\renewcommand{\arraystretch}{1.5}
\resizebox{150mm}{55mm}{ 
\begin{tabular}{|c|c|c|c|c|c|c|c|c|c|c|c|} \hline
\multirow{5}{*}{origin} &{Dataset} &{method} &{Smeasure} &{wFmeasure} &{MAE}  &{adpEm}  &{meanEm} &{maxEm} &{adpFm}    &{meanFm} &{maxFm} \\  \cline{2-12}
& {CAMO} & {UR-SINetv2} & {0.814} & {0.758}  & 0.067 & 0.894 & 0.891 & 0.895  & 0.796 & 0.795 & 0.802    \\  \cline{2-12}
& {CHAMELEON} & {UR-SINetv2} & 0.901 & 0.862	& 0.023 	& 0.962 & 0.96 & 0.965  & 0.872 & 0.873  & 0.883  \\ \cline{2-12}
& {COD10K} & {UR-SINetv2} & 0.816 & 0.708	& 0.033 	& 0.901 & 0.903 & 0.907  & 0.731 & 0.742  & 0.756  \\ \cline{2-12}
& {NC4K} & {UR-SINetv2} & 0.844 & 0.787	& 0.045 	& 0.914 & 0.91 & 0.915  & 0.817 & 0.819  & 0.826  \\ \hline
\multirow{4}{*}{ART} 
& {CAMO} & {UR-SINetv2} & {0.693} & {0.573}  & 0.131 & 0.784 & 0.777 & 0.79  & 0.629 & 0.623 & 0.633     \\  \cline{2-12}
& {CHAMELEON} & {UR-SINetv2} & 0.778 & 0.665	& 0.073 	& 0.853 & 0.855 & 0.87  & 0.701 & 0.703  & 0.718  \\ \cline{2-12}
& {COD10K} & {UR-SINetv2} & 0.723 & 0.547	& 0.065 	& 0.807 & 0.811 & 0.821  & 0.583  & 0.59  & 0.605  \\ \cline{2-12}
& {NC4K} & {UR-SINetv2} & 0.741 & 0.619	& 0.09 	& 0.821 & 0.818 & 0.824  & 0.663 & 0.665  & 0.675  \\ \hline

\multirow{4}{*}{EDSR} &{CAMO} &{UR-SINetv2} &0.689 &0.569  &0.136  &0.781  &0.773  &0.785  &0.626  &0.62 &0.628 \\ \cline{2-12}

&{CHAMELEON} &{UR-SINetv2} &0.779 &0.663	&0.073 	&0.856 &0.852 &0.859  &0.701 &0.7  &0.71  \\ \cline{2-12}
&{COD10K} &{UR-SINetv2} &0.722 &0.544	&0.066 	&0.805 &0.808 &0.819  &0.58 &0.588  &0.605  \\ \cline{2-12}
&{NC4K} &{UR-SINetv2} &0.739 &0.615	&0.092 	&0.817 &0.815 &0.822  &0.659 &0.66  &0.671  \\ \hline

\multirow{4}{*}{SWINIR} &{CAMO} &{UR-SINetv2} &0.69 &0.568  &0.134  &0.784  &0.778  &0.795  &0.625  &0.619 &0.632 \\ \cline{2-12}

&{CHAMELEON} &{UR-SINetv2} &0.776 &0.658	&0.073 	&0.849 &0.853 &0.87  &0.692 &0.695  &0.711  \\ \cline{2-12}
&{COD10K} &{UR-SINetv2} &0.722 &0.545	&0.066 	&0.804 &0.809 &0.819  &0.58 &0.588  &0.605  \\ \cline{2-12}
&{NC4K} &{UR-SINetv2} &0.738 &0.615	&0.091 	&0.818 &0.815 &0.822  &0.659 &0.66  &0.671  \\ \hline

\multirow{4}{*}{ESRGAN}&{CAMO} &{UR-SINetv2} &0.66 &0.528  &0.138  &0.78  &0.763  &0.768  &0.594  &0.584 &0.588 \\ \cline{2-12}

&{CHAMELEON} &{UR-SINetv2} &0.78 &0.66	&0.067 	&0.864 &0.865 &0.876  &0.699 &0.7  &0.711  \\ \cline{2-12}
&{COD10K} &{UR-SINetv2} &0.718 &0.537	&0.063 	&0.809 &0.814 &0.824  &0.575 &0.583  &0.597  \\ \cline{2-12}
&{NC4K} &{UR-SINetv2} &0.721 &0.589	&0.093 	&0.811 &0.805 &0.809  &0.641 &0.64  &0.646  \\ \hline

\multirow{4}{*}{CAT}& {CAMO} & {UR-SINetv2} & {0.693} & {0.573}  & 0.131 & 0.788 & 0.78 & 0.793  & 0.629 & 0.623 & 0.633    \\  \cline{2-12}
& {CHAMELEON} & {UR-SINetv2} & 0.783 & 0.672	& 0.072 	& 0.855 & 0.856 & 0.868  & 0.706 & 0.708  & 0.722  \\ \cline{2-12}
& {COD10K} & {UR-SINetv2} & 0.723 & 0.547	& 0.065 	& 0.807 & 0.81 & 0.82  & 0.583 & 0.591  & 0.607  \\ \cline{2-12}
& {NC4K} & {UR-SINetv2} & 0.742 & 0.622	& 0.089 	& 0.822 & 0.819 & 0.825  & 0.666 & 0.667  & 0.677  \\ \hline
\end{tabular}
}	
\end{table}

\begin{table}
\center
\caption{COD-SR-C2FNet}
\renewcommand{\arraystretch}{1.5}
\resizebox{150mm}{55mm}
{
\begin{tabular}{|c|c|c|c|c|c|c|c|c|c|c|c|} \hline
\multirow{5}{*}{origin} &{Dataset} &{method} &{Smeasure} &{wFmeasure} &{MAE}  &{adpEm}  &{meanEm} &{maxEm} &{adpFm}    &{meanFm} &{maxFm} \\  \cline{2-12}
& {CAMO} & {C2FNet} & {0.796} & {0.719}  & 0.080 & 0.865 & 0.854 & 0.864  & 0.764 & 0.762 & 0.771     \\  \cline{2-12}
& {CHAMELEON} & {C2FNet} & 0.886 & 0.825	& 0.033 	& 0.93 & 0.933 & 0.944  & 0.833 & 0.841  & 0.86  \\ \cline{2-12}
& {COD10K} & {C2FNet} & 0.813 & 0.686	& 0.036 	& 0.886 & 0.890 & 0.900  & 0.703 & 0.723  & 0.743  \\ \cline{2-12}
& {NC4K} & {C2FNet} & 0.838 & 0.762	& 0.049 	& 0.901 & 0.897 & 0.904  & 0.788 & 0.795  & 0.810    \\ \hline
\multirow{4}{*}{ART} 
& {CAMO} & {C2FNet} & {0.66} & {0.503}  & 0.13 & 0.758 & 0.717 & 0.734  & 0.585 & 0.562 & 0.575     \\  \cline{2-12}
& {CHAMELEON} & {C2FNet} & 0.748 & 0.605	& 0.08 	& 0.839 & 0.817 & 0.832  & 0.658 & 0.651  & 0.66  \\ \cline{2-12}
& {COD10K} & {C2FNet} & 0.701 & 0.5	& 0.069 	& 0.787 & 0.782 & 0.789  & 0.544  & 0.548  & 0.56  \\ \cline{2-12}
& {NC4K} & {C2FNet} &0.726  &0.586 	&0.092 	& 0.81 & 0.796 & 0.803  & 0.643 & 0.639  & 0.646  \\ \hline

\multirow{4}{*}{EDSR} &{CAMO} &{C2FNet} &0.648 &0.482  &0.135  &0.746  &0.709  &0.721  &0.563  &0.537 &0.549 \\ \cline{2-12}

&{CHAMELEON} &{C2FNet} &0.745 &0.599	&0.083 	&0.83 &0.811 &0.82  &0.645 &0.644  &0.659  \\ \cline{2-12}
&{COD10K} &{C2FNet} &0.701 &0.497	&0.07 	&0.783 &0.778 &0.785  &0.539 &0.545  &0.559  \\ \cline{2-12}
&{NC4K} &{C2FNet} &0.725 &0.582	&0.093 	&0.809 &0.793 &0.8  &0.64 &0.636  &0.643  \\ \hline

\multirow{4}{*}{SWINIR} &{CAMO} &{C2FNet} &0.652 &0.491  &0.133  &0.749  &0.709  &0.726  &0.574  &0.549 &0.564 \\ \cline{2-12}

&{CHAMELEON} &{C2FNet} &0.745 &0.597	&0.081 	&0.827 &0.807 &0.814  &0.648 &0.641  &0.653  \\ \cline{2-12}
&{COD10K} &{C2FNet} &0.7 &0.496	&0.071 	&0.783 &0.778 &0.786  &0.539 &0.544  &0.557  \\ \cline{2-12}
&{NC4K} &{C2FNet} &0.722. &0.579	&0.094 	&0.807 &0.792 &0.799  &0.637 &0.633  &0.64  \\ \hline

\multirow{4}{*}{ESRGAN}&{CAMO} &{C2FNet} &0.635 &0.461  &0.135  &0.737  &0.684  &0.718  &0.551  &0.522 &0.531 \\ \cline{2-12}

&{CHAMELEON} &{C2FNet} &0.754 &0.619	&0.072 	&0.857 &0.831 &0.855  &0.677 &0.667  &0.675  \\ \cline{2-12}
&{COD10K} &{C2FNet} &0.7 &0.495	&0.066 	&0.788 &0.783 &0.791  &0.541 &0.546  &0.557  \\ \cline{2-12}
&{NC4K} &{C2FNet} &0.711 &0.563	&0.093 	&0.803 &0.784 &0.794  &0.626 &0.62  &0.626  \\ \hline

\multirow{4}{*}{CAT}& {CAMO} & {C2FNet} & {0.656} & {0.498}  & 0.132 & 0.757 & 0.712 & 0.738  & 0.582 & 0.557 & 0.57    \\  \cline{2-12}
& {CHAMELEON} & {C2FNet} & 0.75 & 0.602	& 0.08 	& 0.836 & 0.817 & 0.833  & 0.655 & 0.647  & 0.66  \\ \cline{2-12}
& {COD10K} & {C2FNet} & 0.701 & 0.499	& 0.069 	& 0.786 & 0.782 & 0.789  & 0.542 & 0.547  & 0.56  \\ \cline{2-12}
& {NC4K} & {C2FNet} & 0.727 & 0.586	& 0.092 	& 0.81 & 0.796 & 0.803  & 0.643 & 0.64& 0.647  \\ \hline
\end{tabular}
}
\end{table}

\begin{table}
\center
\caption{COD-SR-C2FNet-TSCVT}
\renewcommand{\arraystretch}{1.5}
\resizebox{150mm}{55mm}{ 
\begin{tabular}{|c|c|c|c|c|c|c|c|c|c|c|c|} \hline
\multirow{5}{*}{origin} &{Dataset} &{method} &{Smeasure} &{wFmeasure} &{MAE}  &{adpEm}  &{meanEm} &{maxEm} &{adpFm}    &{meanFm} &{maxFm} \\  \cline{2-12}
& {CAMO} & {C2FNet-TSCVT} & {0.799} & {0.730}  & 0.077 & 0.869 & 0.859 & 0.869  & 0.777 & 0.770 & 0.779     \\  \cline{2-12}
& {CHAMELEON} & {C2FNet-TSCVT} & 0.893 & 0.845	& 0.028 	& 0.948 & 0.947 & 0.958  & 0.856 & 0.857  & 0.877  \\ \cline{2-12}
& {COD10K} & {C2FNet-TSCVT} & 0.811 & 0.691	& 0.036 	& 0.890 & 0.887 & 0.896  & 0.718 & 0.725  & 0.742  \\ \cline{2-12}
& {NC4K} & {C2FNet-TSCVT} & 0.840 & 0.770	& 0.048 	& 0.900 & 0.896 & 0.904  & 0.799 & 0.802  & 0.814  \\ \hline
\multirow{4}{*}{ART} 
& {CAMO} & {C2FNet-TSCVT} & {0.652} & {0.492}  & 0.136 & 0.747 & 0.72 & 0.735  & 0.554 & 0.542 & 0.554     \\  \cline{2-12}
& {CHAMELEON} & {C2FNet-TSCVT} & 0.733 & 0.589	& 0.087 	& 0.83 & 0.805 & 0.814  & 0.644 & 0.634  & 0.642  \\ \cline{2-12}
& {COD10K} & {C2FNet-TSCVT} & 0.696 & 0.496	& 0.076 	& 0.777 & 0.774 & 0.787  & 0.54  & 0.543  & 0.557  \\ \cline{2-12}
& {NC4K} & {C2FNet-TSCVT} &0.719  &0.577 	& 0.098 	& 0.8 & 0.789 & 0.796  & 0.63 & 0.628  & 0.636  \\ \hline

\multirow{4}{*}{EDSR} &{CAMO} &{C2FNet-TSCVT} &0.643 &0.478  &0.14  &0.736  &0.707  &0.722  &0.542  &0.528 &0.536 \\ \cline{2-12}

&{CHAMELEON} &{C2FNet-TSCVT} &0.722 &0.568	&0.092 	&0.809 &0.79 &0.798  &0.625 &0.612  &0.621  \\ \cline{2-12}
&{COD10K} &{C2FNet-TSCVT} &0.693 &0.491	&0.077 	&0.774 &0.769 &0.782  &0.535 &0.537  &0.552  \\ \cline{2-12}
&{NC4K} &{C2FNet-TSCVT} &0.715 &0.57	&0.099 	&0.796 &0.785 &0.792  &0.621 &0.62  &0.628  \\ \hline

\multirow{4}{*}{SWINIR} &{CAMO} &{C2FNet-TSCVT} &0.651 &0.491  &0.141  &0.734  &0.71  &0.723  &0.545  &0.54 &0.547 \\ \cline{2-12}

&{CHAMELEON} &{C2FNet-TSCVT} &0.72 &0.569	&0.093 	&0.813 &0.79 &0.8  &0.63 &0.614  &0.626  \\ \cline{2-12}
&{COD10K} &{C2FNet-TSCVT} &0.694 &0.493	&0.077 	&0.775 &0.773 &0.786  &0.536 &0.539  &0.554  \\ \cline{2-12}
&{NC4K} &{C2FNet-TSCVT} &0.715 &0.57	&0.1 	&0.796 &0.785 &0.792  &0.622 &0.621  &0.629  \\ \hline

\multirow{4}{*}{ESRGAN}&{CAMO} &{C2FNet-TSCVT} &0.626 &0.456  &0.142  &0.725  &0.692  &0.711  &0.53  &0.513 &0.528 \\ \cline{2-12}

&{CHAMELEON} &{C2FNet-TSCVT} &0.733 &0.59	&0.082 	&0.826 &0.814 &0.822  &0.645 &0.639  &0.649  \\ \cline{2-12}
&{COD10K} &{C2FNet-TSCVT} &0.689 &0.485	&0.072 	&0.78 &0.774 &0.784  &0.534 &0.535  &0.546  \\ \cline{2-12}
&{NC4K} &{C2FNet-TSCVT} &0.696 &0.543	&0.099 	&0.789 &0.773 &0.783  &0.607 &0.6  &0.606  \\ \hline

\multirow{4}{*}{CAT}& {CAMO} & {C2FNet-TSCVT} & {0.65} & {0.489}  & 0.138 & 0.743 & 0.719 & 0.735  & 0.553 & 0.54 & 0.552    \\  \cline{2-12}
& {CHAMELEON} & {C2FNet-TSCVT} & 0.729 & 0.583	& 0.088 	& 0.816 & 0.796 & 0.804  & 0.638 & 0.629  & 0.636  \\ \cline{2-12}
& {COD10K} & {C2FNet-TSCVT} & 0.697 & 0.497	& 0.075 	& 0.78 & 0.775 & 0.788  & 0.542 & 0.544  & 0.559  \\ \cline{2-12}
& {NC4K} & {C2FNet-TSCVT} & 0.719 & 0.578	& 0.098 	& 0.8 & 0.791 & 0.798  & 0.629 & 0.628& 0.636  \\ \hline
\end{tabular}
}
\end{table}

\begin{table}
\center
\caption{COD-SR-CamoFormer}
\renewcommand{\arraystretch}{1.5}
\resizebox{150mm}{55mm}{ 
\begin{tabular}{|c|c|c|c|c|c|c|c|c|c|c|c|} \hline
\multirow{5}{*}{origin} &{Dataset} &{method} &{Smeasure} &{wFmeasure} &{MAE}  &{adpEm}  &{meanEm} &{maxEm} &{adpFm}    &{meanFm} &{maxFm} \\  \cline{2-12}
& {CAMO} & {CamoFormer} & {0.872} & {0.831}  & 0.046 & 0.931 & 0.929 & 0.938  & 0.853 & 0.854 & 0.868     \\  \cline{2-12}
& {CHAMELEON} & {CamoFormer} & 0.91 & 0.859	& 0.023 	& 0.957 & 0.956 & 0.965  & 0.868 & 0.874  & 0.895  \\ \cline{2-12}
& {COD10K} & {CamoFormer} & 0.869 & 0.786	& 0.023 	& 0.931 & 0.932 & 0.939  & 0.794 & 0.811  & 0.829  \\ \cline{2-12}
& {NC4K} & {CamoFormer} & 0.892 & 0.847	& 0.030 	& 0.941 & 0.939 & 0.946  & 0.863 & 0.868  & 0.880  \\ \hline

\multirow{4}{*}{ART} 
& {CAMO} & {CamoFormer} & {0.766} & {0.661}  & 0.094 & 0.843 & 0.835 & 0.845  & 0.712 & 0.709 & 0.723     \\  \cline{2-12}
& {CHAMELEON} & {CamoFormer} & 0.806 & 0.693	& 0.06 	& 0.872 & 0.867 & 0.88  & 0.73 & 0.732  & 0.752  \\ \cline{2-12}
& {COD10K} & {CamoFormer} & 0.79 & 0.652	& 0.046 	& 0.857 & 0.862 & 0.872  & 0.676  & 0.691  & 0.709  \\ \cline{2-12}
& {NC4K} & {CamoFormer} &0.804  &0.706 	& 0.065 	& 0.865 & 0.864 & 0.872  & 0.742 & 0.745  & 0.759  \\ \hline

\multirow{4}{*}{EDSR} &{CAMO} &{CamoFormer} &0.76 &0.655  &0.099  &0.84  &0.83  &0.84  &0.707  &0.703 &0.715 \\ \cline{2-12}

&{CHAMELEON} &{CamoFormer} &0.809 &0.697	&0.057 	&0.873 &0.875 &0.89  &0.731 &0.735  &0.76  \\ \cline{2-12}
&{COD10K} &{CamoFormer} &0.788 &0.649	&0.047 	&0.853 &0.859 &0.87  &0.673 &0.687  &0.706  \\ \cline{2-12}
&{NC4K} &{CamoFormer} &0.803 &0.704	&0.066 	&0.863 &0.862 &0.871  &0.738 &0.742  &0.757  \\ \hline

\multirow{4}{*}{SWINIR} &{CAMO} &{CamoFormer} &0.761 &0.657  &0.098  &0.843  &0.832  &0.843  &0.709  &0.705 &0.72 \\ \cline{2-12}

&{CHAMELEON} &{CamoFormer} &0.806 &0.693	&0.06 	&0.868 &0.867 &0.884  &0.726 &0.731  &0.754  \\ \cline{2-12}
&{COD10K} &{CamoFormer} &0.788 &0.648	&0.046 	&0.853 &0.861 &0.871  &0.671 &0.686  &0.706  \\ \cline{2-12}
&{NC4K} &{CamoFormer} &0.802 &0.702	&0.067 	&0.862 &0.861 &0.869  &0.738 &0.741  &0.756  \\ \hline

\multirow{4}{*}{ESRGAN}&{CAMO} &{CamoFormer} &0.725 &0.607  &0.107  &0.82  &0.804  &0.815  &0.669  &0.661 &0.669 \\ \cline{2-12}

&{CHAMELEON} &{CamoFormer} &0.798 &0.685	&0.059 	&0.886 &0.877 &0.889  &0.726 &0.729  &0.747  \\ \cline{2-12}
&{COD10K} &{CamoFormer} &0.779 &0.634	&0.046 	&0.853 &0.857 &0.866  &0.662 &0.675  &0.693  \\ \cline{2-12}
&{NC4K} &{CamoFormer} &0.785 &0.677	&0.07 	&0.853 &0.85 &0.859  &0.717 &0.72  &0.734  \\ \hline

\multirow{4}{*}{CAT}& {CAMO} & {CamoFormer} & {0.764} & {0.66}  & 0.096 & 0.842 & 0.833 & 0.843  & 0.712 & 0.708 & 0.723    \\  \cline{2-12}
& {CHAMELEON} & {CamoFormer} & 0.809 & 0.696	& 0.059 	& 0.871 & 0.87 & 0.883  & 0.73 & 0.735  & 0.756  \\ \cline{2-12}
& {COD10K} & {CamoFormer} & 0.791 & 0.653	& 0.045 	& 0.858 & 0.863 & 0.872  & 0.678 & 0.692  & 0.71  \\ \cline{2-12}
& {NC4K} & {CamoFormer} & 0.805 & 0.708	& 0.065 	& 0.866 & 0.865 & 0.873  & 0.743 & 0.747& 0.761  \\ \hline
\end{tabular}
	}
\end{table}

\begin{table}
\center
\caption{COD-SR-DGNet}
\renewcommand{\arraystretch}{1.5}
\resizebox{150mm}{55mm}{ 
\begin{tabular}{|c|c|c|c|c|c|c|c|c|c|c|c|} \hline
\multirow{5}{*}{origin} &{Dataset} &{method} &{Smeasure} &{wFmeasure} &{MAE}  &{adpEm}  &{meanEm} &{maxEm} &{adpFm}    &{meanFm} &{maxFm} \\  \cline{2-12}
& {CAMO} & {DGNet} & {0.839} & {0.769}  & 0.057 & 0.906 & 0.901 & 0.915  & 0.804 & 0.806 & 0.822     \\  \cline{2-12}
& {CHAMELEON} & {DGNet} & 0.89 & 0.816	& 0.029 	& 0.935 & 0.938 & 0.956  & 0.822 & 0.834  & 0.865  \\ \cline{2-12}
& {COD10K} & {DGNet} & 0.822 & 0.693	& 0.033 	& 0.879 & 0.896 & 0.911  & 0.698 & 0.728  & 0.759  \\ \cline{2-12}
& {NC4K} & {DGNet} & 0.857 & 0.784	& 0.042 	& 0.910 & 0.911 & 0.922  & 0.803 & 0.814  & 0.833  \\ \hline
\multirow{4}{*}{ART} 
& {CAMO} & {DGNet} & {0.738} & {0.607}  & 0.101 & 0.829 & 0.799 & 0.816  & 0.678 & 0.661 & 0.672     \\  \cline{2-12}
& {CHAMELEON} & {DGNet} & 0.821 & 0.697	& 0.051 	& 0.886 & 0.891 & 0.914  & 0.72 & 0.73  & 0.764  \\ \cline{2-12}
& {COD10K} & {DGNet} & 0.763 & 0.589	& 0.05 	& 0.821 & 0.839 & 0.858  & 0.61  & 0.635  & 0.665  \\ \cline{2-12}
& {NC4K} & {DGNet} &0.786  &0.667 	& 0.069 	& 0.853 & 0.85 & 0.863  & 0.708 & 0.714  & 0.731  \\ \hline

\multirow{4}{*}{EDSR} &{CAMO} &{DGNet} &0.738 &0.608  &0.103  &0.828  &0.801  &0.818  &0.677  &0.662 &0.674 \\ \cline{2-12}

&{CHAMELEON} &{DGNet} &0.822 &0.698	&0.051 	&0.883 &0.886 &0.91  &0.719 &0.731  &0.767  \\ \cline{2-12}
&{COD10K} &{DGNet} &0.762 &0.588	&0.05 	&0.82 &0.839 &0.857  &0.609 &0.633  &0.664  \\ \cline{2-12}
&{NC4K} &{DGNet} &0.785 &0.665	&0.07 	&0.851 &0.849 &0.862  &0.706 &0.712  &0.73  \\ \hline

\multirow{4}{*}{SWINIR} &{CAMO} &{DGNet} &0.736 &0.604  &0.103  &0.826  &0.799  &0.819  &0.675  &0.659 &0.67 \\ \cline{2-12}

&{CHAMELEON} &{DGNet} &0.821 &0.696	&0.051 	&0.885 &0.887 &0.911  &0.72 &0.729  &0.766  \\ \cline{2-12}
&{COD10K} &{DGNet} &0.761 &0.586	&0.05 	&0.819 &0.837 &0.856  &0.608 &0.632  &0.662  \\ \cline{2-12}
&{NC4K} &{DGNet} &0.784 &0.664	&0.07 	&0.85 &0.849 &0.862  &0.705 &0.711  &0.729  \\ \hline

\multirow{4}{*}{ESRGAN}&{CAMO} &{DGNet} &0.725 &0.591  &0.109  &0.816  &0.791  &0.807  &0.662  &0.65 &0.663 \\ \cline{2-12}

&{CHAMELEON} &{DGNet} &0.818 &0.695	&0.052 	&0.876 &0.877 &0.903  &0.719 &0.727  &0.761  \\ \cline{2-12}
&{COD10K} &{DGNet} &0.758 &0.581	&0.051 	&0.818 &0.837 &0.857  &0.603 &0.627  &0.658  \\ \cline{2-12}
&{NC4K} &{DGNet} &0.778 &0.655	&0.072 	&0.848 &0.846 &0.859  &0.697 &0.703  &0.721  \\ \hline

\multirow{4}{*}{CAT}& {CAMO} & {DGNet} & {0.736} & {0.605}  & 0.102 & 0.827 & 0.797 & 0.815  & 0.676 & 0.659 & 0.67    \\  \cline{2-12}
& {CHAMELEON} & {DGNet} & 0.822 & 0.698	& 0.051 	& 0.884 & 0.889 & 0.911  & 0.719 & 0.731  & 0.768  \\ \cline{2-12}
& {COD10K} & {DGNet} & 0.763 & 0.588	& 0.05 	& 0.82 & 0.839 & 0.857  & 0.61 & 0.634  & 0.664  \\ \cline{2-12}
& {NC4K} & {DGNet} & 0.786 & 0.667	& 0.069 	& 0.853 & 0.85 & 0.863  & 0.709 & 0.714& 0.731  \\ \hline
\end{tabular}
	}
\end{table}

\begin{table}
\center
\caption{COD-SR-FAPNet}
\renewcommand{\arraystretch}{1.5}
\resizebox{150mm}{55mm}{ 
\begin{tabular}{|c|c|c|c|c|c|c|c|c|c|c|c|} \hline
\multirow{5}{*}{origin} &{Dataset} &{method} &{Smeasure} &{wFmeasure} &{MAE}  &{adpEm}  &{meanEm} &{maxEm} &{adpFm}    &{meanFm} &{maxFm} \\  \cline{2-12}
& {CAMO} & {FAPNet} & {0.815} & {0.734}  & 0.076 & 0.877 & 0.865 & 0.880  & 0.776 & 0.776 & 0.792     \\  \cline{2-12}
& {CHAMELEON} & {FAPNet} &0.893      &0.825   &0.028 &0.925 &0.94  &0.956 &0.827 &0.842  &0.869   \\ \cline{2-12}
& {COD10K} & {FAPNet} & 0.822 & 0.694	& 0.036 	& 0.875 & 0.888 & 0.902  & 0.707 & 0.731  & 0.758  \\ \cline{2-12}
& {NC4K} & {FAPNet} & 0.851 & 0.775	& 0.047 	& 0.903 & 0.899 & 0.910  & 0.804 & 0.810  & 0.826  \\ \hline
\multirow{4}{*}{ART} 
& {CAMO} & {FAPNet} & {0.704} & {0.551}  & 0.13 & 0.791 & 0.761 & 0.781  & 0.631 & 0.608 & 0.624     \\  \cline{2-12}
& {CHAMELEON} & {FAPNet} & 0.793 & 0.644	& 0.071 	& 0.856 & 0.847 & 0.879  & 0.686 & 0.688  & 0.732  \\ \cline{2-12}
& {COD10K} & {FAPNet} & 0.732 & 0.527	& 0.07 	& 0.771 & 0.789 & 0.817  & 0.557  & 0.578  & 0.615  \\ \cline{2-12}
& {NC4K} & {FAPNet} &0.753  &0.603 	& 0.092 	& 0.813 & 0.805 & 0.824  & 0.654 & 0.656  & 0.679  \\ \hline

\multirow{4}{*}{EDSR} &{CAMO} &{FAPNet} &0.706 &0.551  &0.132  &0.789  &0.761  &0.782  &0.627  &0.608 &0.625 \\ \cline{2-12}

&{CHAMELEON} &{FAPNet} &0.793 &0.64	&0.074 	&0.847 &0.842 &0.878  &0.68 &0.684  &0.727  \\ \cline{2-12}
&{COD10K} &{FAPNet} &0.731 &0.525	&0.071 	&0.769 &0.787 &0.819  &0.556 &0.576  &0.615  \\ \cline{2-12}
&{NC4K} &{BGNet} &0.75 &0.597	&0.093 	&0.81 &0.802 &0.823  &0.649 &0.651  &0.676  \\ \hline

\multirow{4}{*}{SWINIR} &{CAMO} &{FAPNet} &0.706 &0.553  &0.13  &0.796  &0.765  &0.787  &0.629  &0.61 &0.628 \\ \cline{2-12}

&{CHAMELEON} &{FAPNet} &0.793 &0.638	&0.072 	&0.849 &0.836 &0.865  &0.684 &0.684  &0.721  \\ \cline{2-12}
&{COD10K} &{FAPNet} &0.73 &0.523	&0.072 	&0.768 &0.785 &0.816  &0.554 &0.574  &0.611  \\ \cline{2-12}
&{NC4K} &{FAPNet} &0.749 &0.596	&0.093 	&0.808 &0.801 &0.821  &0.648 &0.649  &0.674  \\ \hline

\multirow{4}{*}{ESRGAN}&{CAMO} &{FAPNet} &0.679 &0.509  &0.136  &0.77  &0.733  &0.757  &0.593  &0.571 &0.588 \\ \cline{2-12}

&{CHAMELEON} &{FAPNet} &0.783 &0.628	&0.077 	&0.839 &0.836 &0.871  &0.679 &0.675  &0.702  \\ \cline{2-12}
&{COD10K} &{FAPNet} &0.723 &0.513	&0.072 	&0.769 &0.785 &0.817  &0.547 &0.565  &0.6  \\ \cline{2-12}
&{NC4K} &{FAPNet} &0.732 &0.572	&0.097 	&0.803 &0.789 &0.809  &0.633 &0.63  &0.649  \\ \hline

\multirow{4}{*}{CAT}& {CAMO} & {FAPNet} & {0.711} & {0.561}  & 0.127 & 0.799 & 0.771 & 0.794  & 0.636 & 0.618 & 0.635    \\  \cline{2-12}
& {CHAMELEON} & {FAPNet} & 0.8 & 0.656	& 0.069 	& 0.856 & 0.85 & 0.882  & 0.696 & 0.7  & 0.74  \\ \cline{2-12}
& {COD10K} & {FAPNet} & 0.732 & 0.528	& 0.07 	& 0.772 & 0.789 & 0.818  & 0.559 & 0.58  & 0.616  \\ \cline{2-12}
& {NC4K} & {FAPNet} & 0.754 & 0.605	& 0.091 	& 0.814 & 0.806 & 0.826  & 0.657 & 0.658 & 0.682  \\ \hline
\end{tabular}
	}
\end{table}

\begin{table}
\center
\caption{COD-SR-Sinet}
\renewcommand{\arraystretch}{1.5}
\resizebox{150mm}{55mm}{ 
\begin{tabular}{|c|c|c|c|c|c|c|c|c|c|c|c|} \hline
\multirow{5}{*}{origin} &{Dataset} &{method} &{Smeasure} &{wFmeasure} &{MAE}  &{adpEm}  &{meanEm} &{maxEm} &{adpFm}    &{meanFm} &{maxFm} \\  \cline{2-12}
& {CAMO} & {Sinet} & {0.745} & {0.644}  & 0.092 & 0.825 & 0.804 & 0.829  & 0.712 & 0.702 & 0.708     \\  \cline{2-12}
& {CHAMELEON} & {Sinet} & 0.872 & 0.806	& 0.034 	& 0.938 & 0.936 & 0.946  & 0.823 & 0.827  & 0.838  \\ \cline{2-12}
& {COD10K} & {Sinet} & 0.776 & 0.631	& 0.043 	& 0.867 & 0.864 & 0.874  & 0.667 & 0.679  & 0.691  \\ \cline{2-12}
& {NC4K} & {Sinet} & 0.808 & 0.723	& 0.058 	& 0.883 & 0.871 & 0.883  & 0.768 & 0.769  & 0.775  \\ \hline
\multirow{4}{*}{ART} 
& {CAMO} & {Sinet} & {0.64} & {0.463}  & 0.13 & 0.73 & 0.682 & 0.714  & 0.557 & 0.524 & 0.536     \\  \cline{2-12}
& {CHAMELEON} & {Sinet} & 0.752 & 0.612	& 0.072 	& 0.847 & 0.816 & 0.829  & 0.673 & 0.661  & 0.667  \\ \cline{2-12}
& {COD10K} & {Sinet} & 0.708 & 0.513	& 0.061 	& 0.805 & 0.796 & 0.803  & 0.563  & 0.566  & 0.575  \\ \cline{2-12}
& {NC4K} & {Sinet} &0.724  &0.586 	& 0.088 	& 0.813 & 0.794 & 0.804  & 0.65 & 0.643  & 0.648  \\ \hline

\multirow{4}{*}{EDSR} &{CAMO} &{Sinet} &0.636 &0.458  &0.133  &0.724  &0.682  &0.709  &0.548  &0.518 &0.527 \\ \cline{2-12}

&{CHAMELEON} &{Sinet} &0.747 &0.604	&0.076 	&0.833 &0.81 &0.817  &0.663 &0.652  &0.658  \\ \cline{2-12}
&{COD10K} &{Sinet} &0.708 &0.513	&0.062 	&0.803 &0.794 &0.8  &0.563 &0.566  &0.575  \\ \cline{2-12}
&{NC4K} &{BGNet} &0.724 &0.586	&0.088 	&0.815 &0.794 &0.805  &0.65 &0.643  &0.648  \\ \hline

\multirow{4}{*}{SWINIR} &{CAMO} &{Sinet} &0.638 &0.461  &0.131  &0.729  &0.684  &0.716  &0.556  &0.522 &0.535 \\ \cline{2-12}

&{CHAMELEON} &{Sinet} &0.746 &0.599	&0.077 	&0.827 &0.815 &0.824  &0.652 &0.646  &0.654  \\ \cline{2-12}
&{COD10K} &{Sinet} &0.708 &0.514	&0.062 	&0.802 &0.795 &0.801  &0.562 &0.567  &0.576  \\ \cline{2-12}
&{NC4K} &{Sinet} &0.724 &0.586	&0.089 	&0.812 &0.795 &0.804  &0.649 &0.643  &0.647  \\ \hline

\multirow{4}{*}{ESRGAN}&{CAMO} &{Sinet} &0.622 &0.437  &0.134  &0.717  &0.67  &0.715  &0.529  &0.499 &0.516 \\ \cline{2-12}

&{CHAMELEON} &{Sinet} &0.728 &0.577	&0.079 	&0.809 &0.795 &0.802  &0.631 &0.624  &0.63  \\ \cline{2-12}
&{COD10K} &{Sinet} &0.7 &0.5	&0.063 	&0.8 &0.787 &0.795  &0.551 &0.554  &0.561  \\ \cline{2-12}
&{NC4K} &{Sinet} &0.71 &0.567	&0.09 	&0.807 &0.784 &0.8  &0.639 &0.628  &0.633  \\ \hline

\multirow{4}{*}{CAT}& {CAMO} & {Sinet} & {0.643} & {0.471}  & 0.129 & 0.74 & 0.696 & 0.727  & 0.56 & 0.531 & 0.541    \\  \cline{2-12}
& {CHAMELEON} & {Sinet} & 0.756 & 0.621	& 0.071 	& 0.844 & 0.816& 0.829  & 0.682 & 0.671  & 0.678  \\ \cline{2-12}
& {COD10K} & {Sinet} & 0.708 & 0.514	& 0.061 	& 0.805 & 0.796 & 0.802  & 0.565 & 0.567  & 0.575  \\ \cline{2-12}
& {NC4K} & {Sinet} & 0.726 & 0.59	& 0.087 	& 0.815 & 0.796 & 0.807  & 0.654 & 0.647 & 0.652  \\ \hline
\end{tabular}
	}
\end{table}

\begin{table}
\center
\caption{COD-SINet-v2}
\renewcommand{\arraystretch}{1.5}
\resizebox{150mm}{55mm}{ 
\begin{tabular}{|c|c|c|c|c|c|c|c|c|c|c|c|} \hline
\multirow{5}{*}{origin} &{Dataset} &{method} &{Smeasure} &{wFmeasure} &{MAE}  &{adpEm}  &{meanEm} &{maxEm} &{adpFm}    &{meanFm} &{maxFm} \\  \cline{2-12}
& {CAMO} & {SINet-v2} & {0.820} & {0.743}  & 0.070 & 0.884 & 0.882 & 0.895  & 0.779 & 0.782 & 0.801     \\  \cline{2-12}
& {CHAMELEON} & {SINet-v2} & 0.888 & 0.816	& 0.03 	& 0.929 & 0.941 & 0.961  & 0.816 & 0.835  & 0.867  \\ \cline{2-12}
& {COD10K} & {SINet-v2} & 0.815 & 0.680	& 0.037 	& 0.864 & 0.887 & 0.906  & 0.682 & 0.718  & 0.752  \\ \cline{2-12}
& {NC4K} & {SINet-v2} & 0.847 & 0.770	& 0.048 	& 0.901 & 0.903 & 0.914  & 0.792 & 0.805  & 0.823  \\ \hline
\multirow{4}{*}{ART} 
& {CAMO} & {SINet-v2} & {0.702} & {0.558}  & 0.14 & 0.782 & 0.761 & 0.789  & 0.623 & 0.613 & 0.635     \\  \cline{2-12}
& {CHAMELEON} & {SINet-v2} & 0.778 & 0.634	& 0.082 	& 0.827 & 0.833 & 0.87  & 0.67 & 0.679  & 0.72  \\ \cline{2-12}
& {COD10K} & {SINet-v2} & 0.725 & 0.524	& 0.072 	& 0.765 & 0.791 & 0.82  & 0.547  & 0.574  & 0.609  \\ \cline{2-12}
& {NC4K} & {SINet-v2} &0.746  &0.601 	& 0.096 	& 0.804 & 0.805 & 0.824  & 0.644 & 0.651  & 0.678  \\ \hline

\multirow{4}{*}{EDSR} &{CAMO} &{SINet-v2} &0.698 &0.552  &0.146  &0.775  &0.756  &0.786  &0.614  &0.608 &0.631 \\ \cline{2-12}

&{CHAMELEON} &{SINet-v2} &0.778 &0.632	&0.083 	&0.831 &0.831 &0.858  &0.671 &0.675  &0.706  \\ \cline{2-12}
&{COD10K} &{SINet-v2} &0.723 &0.522	&0.072 	&0.762 &0.789 &0.818  &0.544 &0.571  &0.607  \\ \cline{2-12}
&{NC4K} &{SINet-v2} &0.744 &0.597	&0.099 	&0.8 &0.801 &0.823  &0.64 &0.646  &0.675  \\ \hline

\multirow{4}{*}{SWINIR} &{CAMO} &{SINet-v2} &0.698 &0.552  &0.144  &0.777  &0.76  &0.793  &0.615  &0.608 &0.634 \\ \cline{2-12}

&{CHAMELEON} &{SINet-v2} &0.775 &0.627	&0.083 	&0.824 &0.831 &0.87  &0.663 &0.672  &0.717  \\ \cline{2-12}
&{COD10K} &{SINet-v2} &0.723 &0.522	&0.072 	&0.761 &0.789 &0.819  &0.544 &0.571  &0.608  \\ \cline{2-12}
&{NC4K} &{SINet-v2} &0.744 &0.597	&0.098 	&0.8 &0.802 &0.822  &0.64 &0.647  &0.674  \\ \hline

\multirow{4}{*}{ESRGAN}&{CAMO} &{SINet-v2} &0.672 &0.518  &0.146  &0.777  &0.751  &0.771  &0.591  &0.577 &0.589 \\ \cline{2-12}

&{CHAMELEON} &{SINet-v2} &0.78 &0.63	&0.075 	&0.831 &0.842 &0.873  &0.668 &0.677  &0.709  \\ \cline{2-12}
&{COD10K} &{SINet-v2} &0.72 &0.514	&0.07 	&0.766 &0.792 &0.822  &0.54 &0.566  &0.6  \\ \cline{2-12}
&{NC4K} &{SINet-v2} &0.729 &0.573	&0.099 	&0.798 &0.793 &0.809  &0.625 &0.628  &0.647  \\ \hline

\multirow{4}{*}{CAT}& {CAMO} & {SINet-v2} & {0.702} & {0.558}  & 0.141 & 0.781 & 0.764 & 0.793  & 0.621 & 0.613 & 0.637    \\  \cline{2-12}
& {CHAMELEON} & {SINet-v2} & 0.781 & 0.641	& 0.08 	& 0.827 & 0.835& 0.868  & 0.676 & 0.683  & 0.722  \\ \cline{2-12}
& {COD10K} & {SINet-v2} & 0.725 & 0.525	& 0.072 	& 0.765 & 0.791 & 0.82  & 0.548 & 0.574  & 0.61  \\ \cline{2-12}
& {NC4K} & {SINet-v2} & 0.748 & 0.604	& 0.096 	& 0.804 & 0.806 & 0.826  & 0.646 & 0.653 & 0.681  \\ \hline
\end{tabular}
	}
\end{table}


\end{document}